\documentclass[pdflatex,sn-mathphys-num]{sn-jnl}% Math and Physical Sciences Numbered Reference Style
\usepackage{graphicx}%
\usepackage{multirow}%
\usepackage{amsmath,amssymb,amsfonts}%
\usepackage{amsthm}%
\usepackage{mathrsfs}%
\usepackage[title]{appendix}%
\usepackage{xcolor}%
\usepackage{textcomp}%
\usepackage{manyfoot}%
\usepackage{algorithm}%
\usepackage{algorithmicx}%
\usepackage[hypcap=false]{caption}
\usepackage{graphicx}
\usepackage{caption}
\usepackage{algpseudocode}%
\usepackage{booktabs}%
\usepackage{array}
\usepackage{float}
\usepackage{caption}
\usepackage{xcolor}
\definecolor{newcolor}{rgb}{.8,.349,.1}
\usepackage[draft]{changes}
\setdeletedmarkup{\color{red}{\sout{#1}}}
\setaddedmarkup{\color{red}{#1}}
\usepackage{xcolor}
\usepackage{soul}
\usepackage{listings}%
\theoremstyle{thmstyleone}%
\theoremstyle{thmstyletwo}%

\theoremstyle{thmstylethree}%

\begin{document}
\def\orcidlink#1{}

\title[Article Title]{Impact Detection in Fall Events: Leveraging Spatio-Temporal Graph Convolutional Networks and Recurrent Neural Networks Using 3D Skeletons Data}


\author*[1,4]{\fnm{Tresor Y.} \sur{Koffi}}\email{ytkoffi@cesi.fr}
\author[1]{\fnm{Youssef} \sur{Mourchid}}\email{ymourchid@cesi.fr}
\author[2]{\fnm{Mohammed} \sur{Hindawi}}\email{mhindawi@cesi.fr}
\author[3]{\fnm{Yohan} \sur{Dupuis}}\email{ydupuis@cesi.fr}

\affil*[1]{\orgdiv{CESI}, \orgname{CESI LINEACT}, \orgaddress{\city{Dijon}, \country{France}}}
\affil[2]{\orgdiv{CESI}, \orgname{CESI LINEACT}, \orgaddress{\city{Lyon}, \country{France}}}
\affil[3]{\orgdiv{CESI}, \orgname{CESI LINEACT}, \orgaddress{\city{Paris}, \country{France}}}
\affil[4]{\orgdiv{ED 432}, \orgname{ENSAM}, \orgaddress{\city{Paris}, \country{France}}}

\abstract{Fall represents a significant risk of accidental death among individuals aged over 65, presenting a global health concern. A fall is defined as any event where a person loses balance and moves to an off-position, which may or may not result in an impact where the person hits the ground. While fall detection systems have achieved good results in general, impact detection within falls remains challenging. This study proposes an efficient methodology for accurately detecting impacts within fall events by incorporating 3D joints skeleton data treated as a graph using Spatio-Temporal Graph Convolutional Networks (STGCN), Gated Recurrent Unit (GRU), and Bidirectional Long Short-Term Memory (BiLSTM) layers. By pinpointing impact moments, our approach enhances precision by distinguishing between false falls and actual impacts, contributing to better healthcare resource allocation. Our methodology, evaluated using the improved 3D skeletons UP-Fall dataset, achieves accuracy exceeding 90\% across various fall scenarios. We have made this improved dataset publicly available at \url{https://zenodo.org/records/12773013} to facilitate further research. 
}
\keywords{Impact Detection, Joint Skeleton, Graph Convolution Network, UP-Fall Dataset, Improved 3D Skeletons Data, Healthcare}

\maketitle

\section{Introduction}\label{sec1}

The world is currently grappling a high challenge of addressing the need to protect elderly people from falls. In fact, it is projected that by 2050, one in six individuals globally will be 65 years old or older \citep{mitchell2020global}. Moreover, fall has been identified as a significant cause of accidental deaths, especially among older adults with cognitive impairments \citep{bourke2007evaluation}.
\begin{figure*}
\centering
\includegraphics[width=13cm]{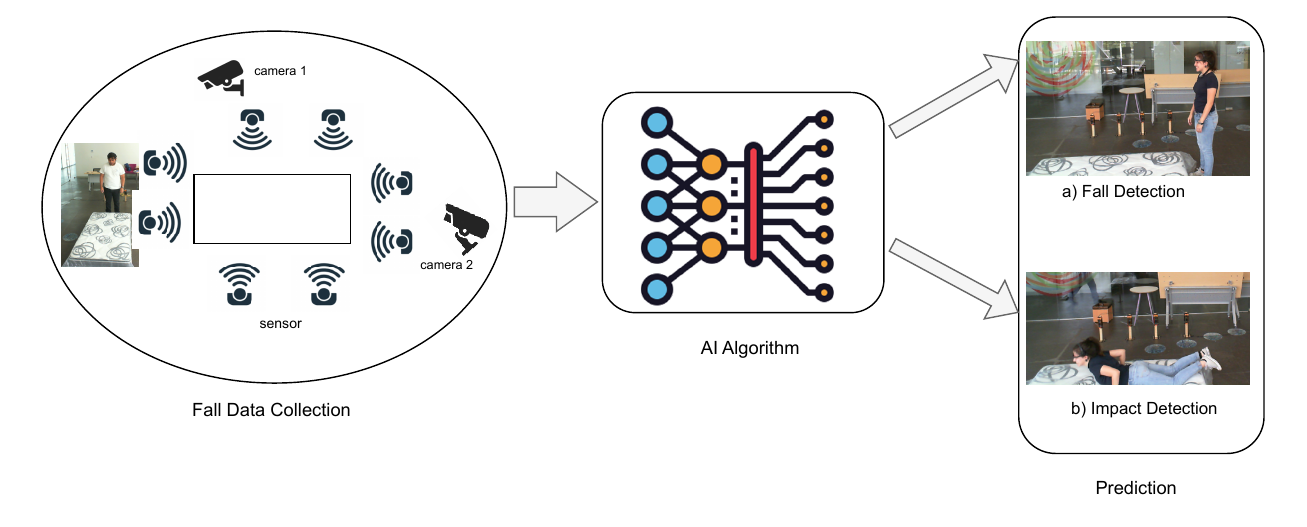}

\caption{a) Fall detected by other algorithms versus b) impact fall detected by our proposed algorithm}
\label{FDS}
\end{figure*}
Indeed, fall is responsible for approximately 646,000 deaths worldwide per year, with elderly individuals being the most vulnerable \citep{ximenes2024impact}. Those having mild cognitive impairment diseases face a higher risk of falling. Given the insufficience of medical care personnel and resources, precise detection of impact when the person hits the ground within a fall event is crucial for effective allocation of healthcare resources. Studies have shown that immediate assistance can significantly reduce and avoid the risk of death and long-term hospitalization. While existing fall detection systems achieve high accuracy in identifying fall events, they fundamentally lack the temporal precision required for accurate impact detection. The critical distinction lies between detecting that a fall has occurred versus pinpointing the exact moment when an individual makes contact with the ground. This temporal precision is essential for reducing false positive rates and improving system reliability. Accurate impact timing enables the differentiation between near-falls, stumbles, and actual ground contact, significantly reducing unnecessary alerts and preventing system overload. Current fall detection methods often trigger alerts during the falling motion rather than at the actual impact moment, leading to higher false positive rates. Therefore, precise impact moment detection provides critical temporal information that enhances the overall accuracy and reduces computational overhead by focusing analysis on the specific moment of interest rather than entire fall sequences. To achieve such precise impact detection, various technologies are employed to collect comprehensive data for fall analysis. Several technologies can be used to collect data for fall detection, including sensors, cameras, and infrared arrays. Sensors, such as accelerometers, gyroscopes, and magnetometers, are typically worn on various parts of the body to monitor motion as outlined by \citep{noury2007fall}, and \citep{mubashir2013survey}. These sensors continuously monitor orientation by capturing raw data on acceleration, angular velocity, and magnetic field strength, which is essential for analyzing movement pattern changes that may indicate a fall. Cameras are another vital technology used for fall detection. They capture visual information, including body posture and movement, providing valuable insights into the dynamics of a fall event \citep{wang2023fall}. Infrared arrays, as an another alternative, detect heat signatures emitted by the human body. These arrays can be particularly useful in low-light conditions or environments where visual information may be limited. By detecting changes in heat distribution, infrared arrays can supply additional information to enhance the accuracy of fall detection systems\citep{yang2022fall}. These technologies complement each other, providing a comprehensive dataset for accurate fall detection. Several approaches are used to detect fall and reduce its consequences.  These approaches can roughly be categorized into:  threshold-based approaches, machine learning-based approaches, and deep learning-based approaches  \citep{wangfusion} . In threshold-based approaches, sensors and infrared techniques are extensively used to identify falls. These methods involve setting predefined thresholds for certain parameters, such as acceleration or heat signatures, and triggering an alarm when these thresholds are exceeded, indicating a potential fall event.
Machine learning-based approaches use multimodal data including image data from cameras, inertial data from accelerometers \citep{ren2019research} and ML models to detect fall, providing a non-intrusive solution for fall detection. These methods are used to analyze human tracking applications and user recognition systems\citep{kottari2019real,espinosa2020application}. Deep learning approaches leverage advanced neural network architectures to analyze sensor data or images, including 3D information from depth-sensing devices like Kinect \citep{kottari2019real} \citep{espinosa2020application}, for the detection of falls.
Among these architectures, Temporal Convolutional Network (TCN) has emerged as a powerful tool for processing sequential data. TCNs excel at capturing long-range temporal dependencies, making them particularly effective for analyzing the temporal aspects of fall events. Unlike traditional recurrent networks, TCNs can process multiple time steps in parallel, leading to more efficient training and inference \citep{ma2024application}. Additionally, recent fall detection systems benefit from attention mechanism \citep{kibet2024sudden}. Attention vectors help the model focus on the most relevant parts of the input sequence, whether it's specific body joints or particular time segments during a potential fall event. This selective focus is especially important in fall detection, as it allows the system to distinguish between normal activities and actual falls by paying attention to critical motion patterns and temporal relationships.
Moreover, in computer vision, deep learning approaches based on graphs have been widely used \citep{mourchid2016image,lafhel2021movie, cherifi2017complex,mourchid2023d}, particularly for skeleton-based action identification due to the similarities between the human skeleton and a graph. The popularity of deep learning models in fall detection has grown due to their ability to capture 3D information such as human poses or limb positions \citep{chen2020fall}. These methods offer several advantages, such as automatically learning complex patterns and features from the data without the need for manual feature engineering.
Deep learning models can effectively handle both spatial and temporal data, making them suitable for analyzing sensor data from accelerometers, gyroscopes, and magnetometers, as well as images or videos from cameras. However, despite these capabilities, accurately detecting impacts within fall events remains a significant challenge due to the high incidence of false alerts as shown in Figure \ref{FDS}.(a) \citep{islam2020deep}. These false alerts often occur when the system misinterprets non-fall movements or activities as falls.
Hence, the challenge lies in distinguishing between genuine falls and activities that may mimic fall-like motions, such as sudden movements, bending, or rapid posture changes. This issue becomes particularly relevant in environments where individuals are engaged in dynamic activities, leading to high risk of false alerts. Moreover, the consequences of false alerts can be significant, causing unnecessary distress for both individuals being monitored and caregivers. To address these challenges, we propose a deep learning architecture that leverages the Spatio-Temporal Graph Convolutional Network (STGCN) and uses 3D joints skeleton data. This model aims to accurately detect impact events where the person makes contact with the ground within a fall event as shown in Figure \ref{FDS}.(b), significantly reducing false positives triggered by non-fall activities. Hence, for the first enhancement, the Spatio-Temporal Graph Convolutional Network (STGCN) framework using ConvLSTM was retained, but an improvement using Gated Recurrent Unit (GRU) layers was introduced. Unlike traditional ConvLSTM layers, GRUs excel in processing sequential data, including dynamic body joint movements, leading to faster training times and enhanced computational performance suitable for real-time applications. Furthermore, the GRU layer operates on the adjacency matrix of body-joints, which can determine the contribution of each body-joint in the fall and can significantly enhance fall classification by differentiating between various types of falls. Key joint analysis aimed to improve sensitivity and accuracy by highlighting specific patterns associated with different falls, such as forward or backward falls, and allows for better contextual understanding. Building upon the solid foundation established by the STGCN with GRU layers, the model undergoes further enhancement with the integration of a Bi-directional Long Short-Term Memory (BiLSTM) layer. By analyzing the sequence data bidirectionally, this layer extracts rich contextual information that goes beyond simple temporal patterns. This bidirectional processing enables the model to gain more understanding of the temporal dynamics inherent in fall events. Moreover, BiLSTM can handle variable-length sequences, crucial for real-world applications where fall events may exhibit varying speeds. By processing sequences of varying lengths, the BiLSTM can capture the entire temporal context of each fall event, regardless of its duration. This versatility allows the model to adapt  to different scenarios without the need for explicit preprocessing of sequence length. The main contributions of this paper are as follows:

\begin{itemize}
\item The use of an improved 3D skeleton dataset for fall detection, which enhances the quality and reliability of input data for impact fall detection systems. This dataset is made publicly available to facilitate further research and development in the field.

\item The use of STGCN layers to capture both spatial and temporal dependencies; 

\item The GRU layers are added for faster and more efficient analysis of temporal dependencies. They also help in determining the contribution of each body-joint in fall events for fall classification.

\item The Bi-directional Long Short-Term Memory (BiLSTM) layer is used to improve the model's analysis of fall sequences and effectively processes variable-length sequences, ensuring robust performance in diverse real-world scenarios.

The sections of this paper are organized as follows:    
In \hyperref[sec:related_work]{Section 2}, we delve into an exploration of the existing state-of-the-art fall detection methods and articulate the motivation behind our proposed approach. \hyperref[sec:The proposed approach]{Section 3}  comprehensively details of our proposed system. \hyperref[sec:Experimental Results]{Section 4} outlines the experimental setup and presents the results. Ultimately, \hyperref[sec:Conclusion]{Section 5}  encapsulates concluding remarks and outlines potential future research.
\end{itemize}

 \section{Related Works}\label{sec:related_work}

In the ongoing pursuit to advance fall detection systems, diverse methodologies have been explored, contributing to a categorization into three principal groups: threshold-based methods, machine learning-based methods, and deep learning-based methods. However, detecting genuine falls using these approaches is still challenging particularly in accurately identifying the impact within a fall event. The moment of impact is a critical factor in identifying a genuine fall, and decreasing any misinterpretation that can result in inaccurate detection and potentially unnecessary alerts. The following subsequent sections offer a more detailed exploration of fall detection approaches, shedding light on their methodologies and providing insights into their challenges.

\subsection{Threshold-based Approaches}

Threshold-based techniques represent the earliest and most straightforward approach to fall detection, identifying falls by analyzing sensor values against predefined thresholds. The fundamental method, as employed by \citet{bourke2008threshold}, evaluates whether specific handcrafted features exceed predetermined values at particular frames, making it computationally lightweight and suitable for resource-constrained embedded systems and wearable devices.
Early implementations focused on simple acceleration-based thresholds, but researchers quickly recognized the challenge of determining optimal reference values \citep{mekruksavanich2022pre}. The variability in human activities and disparities between experimental and real-world conditions pose significant challenges for accurately detecting impact moments when individuals make contact with the ground during fall events.
To address these limitations, adaptive and personalized thresholding approaches have emerged. \citet{li2009accurate} explored adaptive thresholding based on personal gait patterns and sensor calibration to reduce false positives, while \citet{lim2022application} proposed user-specific threshold tuning through personalized gait profiling. \citet{li2023novel} analyzed peak acceleration patterns across multiple fall directions, revealing the inherent limitations of fixed thresholds when fall kinematics differ significantly.
Vision-based threshold systems have incorporated more sophisticated implementations. \citet{mastorakis2014fall} experimented with dynamic silhouette-based thresholds for video-based fall detection in indoor settings, while \cite{de2022fallCombinedDisplacement} presented an advanced approach combining Gaussian Mixture Models and fuzzy logic with threshold-based decisions. Their method uses normalized instantaneous speed thresholds (Ts = 1.8) and fuzzy logic to classify entire fall events into categories (abrupt fall, normal fall, and ADLs), achieving 96.66\% sensitivity. However, this approach focuses on categorizing fall types rather than detecting the precise moment when an individual hits the ground during a fall event. Such temporal precision is crucial for reducing false alerts and efficiently mobilizing emergency services, as it distinguishes between near-falls and actual ground contact moments.
Recent developments have emphasized sensor fusion strategies. \citet{singh2020sensor} introduced adaptive thresholding combined with sensor fusion for noisy real-world conditions, while \citet{nooruddin2020iot} proposed multi-stage thresholding utilizing gyroscope and accelerometer fusion, showing improvements in fall localization for real-time wearable systems. Despite these advances, threshold-based methods remain fundamentally limited in their ability to precisely identify the moment of ground contact within fall sequences.

\subsection{Machine Learning-based Approaches}

Machine learning (ML) approaches emerged as a response to the limitations of threshold-based methods, offering enhanced accuracy and adaptability \citep{wasi2022machine}. These methods address conventional challenges such as poor camera resolution, undesirable sensor data features, and misclassification issues through sophisticated pattern recognition and feature learning \cite{kausar2022automated}.
Traditional ML implementations have leveraged diverse algorithms including K-Nearest Neighbors (KNN), Support Vector Machines (SVM), Gradient Boosting (GBOOST), Random Forest (RF), Naive Bayes (NB), Logistic Regression (LR), and Decision Trees (DT) \citep{putra2017event} to distinguish falls from Activities of Daily Living (ADL). These approaches typically extract features from accelerometer, gyroscope, and magnetometer data, with some studies reporting accuracy around 99\%. For instance, SVM applied to accelerometer and magnetometer data achieved 99.3\% sensitivity and 96\% specificity in controlled settings \cite{droghini2017combined}. However, these traditional ML approaches suffer from several critical limitations: they require extensive manual feature engineering, perform poorly with high-dimensional data, and most importantly, focus on binary fall/non-fall classification rather than identifying the precise impact moment within fall sequences.
Vision-based ML approaches have shown promise in addressing sensor-related limitations. \citet{kim2020implementation} employed Extreme Learning Machine (ELM) classifiers with depth camera data to distinguish between ADL and fall activities across diverse poses including falling, bending, sitting, squatting, walking, and lying, demonstrating versatility across different sensor modalities. Nevertheless, these vision-based ML methods remain constrained by their reliance on handcrafted visual features and struggle with complex temporal dependencies required for accurate impact moment detection.
Hybrid approaches have emerged to combine different methodological strengths. 
\citet{de2022fallSpatioTemporalFusion} proposed a two-channel system integrating threshold-based classification with K-NN machine learning, using Motion History Images (MHI) and spatial features to achieve 98.6\% accuracy in fall event detection. However, while these hybrid systems improve overall fall detection performance, they inherit the limitations of both constituent approaches and do not address impact timing detection, focusing instead on binary event-level classification rather than temporal precision required for impact localization
Advanced ML systems have incorporated context-awareness and personalization \cite{mir2025machine} in fall events. \citet{miranda2022survey} emphasized personalized ML models that adapt classification thresholds to individual walking styles, while \citet{li2025multidimensional} introduced dynamic time warping and motion segmentation for better alignment of windowed analysis with actual fall phases. Despite these advances, personalized ML systems still require extensive individual calibration data and fail to achieve the temporal precision necessary for accurate impact detection.
Notably, our previous work 
\citet{koffi2023machine} was the first to specifically target impact detection using multisensor feature ranking and ML modeling on the UP-FALL dataset. To the best of our knowledge, this represents the pioneering study explicitly focusing on impact localization rather than general fall classification, as existing methods focus on general fall detection without distinguishing the precise moment of ground contact. This work demonstrated the feasibility of detecting impact moments within fall events and provided a foundation for impact-specific detection methodologies. However, this approach revealed fundamental limitations of ML-based impact detection: exclusive reliance on wearable sensors requiring continuous device usage, challenges with user compliance and comfort, and limited ability to model complex spatio-temporal joint relationships essential for precise impact characterization. These limitations necessitate more sophisticated approaches that can leverage visual data and model complex temporal dependencies without requiring continuous sensor wear.

\subsection{Deep Learning-based Approaches}

Deep learning has revolutionized fall detection by enabling automatic feature learning directly from raw sensor data, eliminating the need for manual feature engineering \citep{shen2019fall, csengul2022deep}. Convolutional Neural Networks (CNNs), Recurrent Neural Networks (RNNs), and their combinations have demonstrated superior performance in distinguishing falls from other activities \citep{csengul2022deep}, with CNN-LSTM fusion proving particularly effective \citep{xu2019cnn}.
Early deep learning implementations focused on general fall detection using various architectural combinations. These models demonstrated proficiency in capturing complex patterns and representations within raw sensor data without explicit feature engineering \citep{balasubramaniam2023modified}, marking a significant advancement over traditional ML approaches. However, these early implementations suffered from critical limitations including binary classification focus, lack of temporal precision for impact localization, and computational intensity unsuitable for real-time deployment.
Recent developments have emphasized sophisticated feature fusion strategies. \citet{almukadi2024deep} proposed the Deep Feature Fusion with Computer Vision for Fall Detection and Classification (DFFCV-FDC) technique, combining features from multiple deep networks including MobileNet, DenseNet, and ResNet. Their approach employs Gaussian filtering for noise reduction and improved pelican optimization for hyperparameter selection, ultimately applying denoising autoencoders for fall classification. While achieving high accuracy in fall identification, this multi-network fusion approach presents significant limitations including excessive computational complexity, primary focus on binary classification rather than temporal analysis, and inability to precisely detect impact moments within fall sequences.
Recognizing the importance of temporal precision, researchers have developed specialized models for pre-impact detection. \citet{kim2019machine} proposed a deep learning architecture specifically for pre-impact identification using accelerometer data, while \citet{yu2020novel} applied recurrent neural networks for predicting imminent falls. These pre-impact detection approaches, while valuable for early warnings, exhibit fundamental limitations as they predict future fall likelihood rather than detecting actual impact occurrence, require sensor-based input limiting deployment flexibility, and cannot precisely pinpoint the frame of ground contact when impact actually occurs.
Recent lightweight implementations have addressed deployment constraints while maintaining temporal focus. The TinyFallNet model \citep{turetta2025lightweight} provides a ConvLSTM variant optimized for pre-impact localization, achieving lead times around 477 ms with acceptable accuracy on older-adult datasets, demonstrating the trade-off between latency and localization precision. Similarly, PreFallKD \citep{chi2023prefallkd} employs CNN-ViT knowledge distillation to create compact yet effective pre-impact models with approximately 551 ms lead time and 92.7\% F1-score, illustrating how transformer architectures can enable impact anticipation in wearable settings. Despite their computational efficiency, these lightweight models remain constrained by inherent trade-offs between model size and accuracy, focus on prediction rather than detection, and limited ability to model complex spatio-temporal relationships crucial for precise impact characterization.
Physics-based deep learning approaches have emerged to leverage domain knowledge. TSFallDetect \citep{qu2024physics} integrates embedded sensors with time-series deep learning models on inertial and pressure data for fall prediction, showing potential for embedded real-world impact detection by incorporating physical principles into model design. However, physics-based approaches are limited by their dependence on sensor availability and placement, computational complexity of integrating physical constraints, and continued focus on prediction rather than precise impact moment detection.
Despite these advances, fundamental limitations persist across all deep learning approaches. Most deep learning models employ event-level supervision and sliding-window labeling, preventing reliable detection of exact impact moments. Additionally, existing deep learning methods fail to adequately model the complex spatio-temporal joint relationships essential for impact detection, lack the temporal precision required for frame-level impact localization, and do not leverage the rich structural information available in human skeletal data. Consequently, accurately and reliably identifying the specific instant of ground contact within fall events remains an open research challenge, highlighting the need for specialized approaches that focus on impact moment detection rather than general fall classification and can effectively model complex joint interactions through sophisticated spatio-temporal architectures.

\section{The Proposed Approach}
\label{sec:The proposed approach}
%%\label{}
In this section, we will elaborate on the problem statement, provide a detailed pipeline of our proposed method as shown in Figure \ref{pipeline}, and highlight the sequential steps involved in our methodology. Through this presentation, we aim to provide a clear understanding of our proposed approach for detecting impact within fall events.

\begin{center}
\includegraphics[scale=0.45]{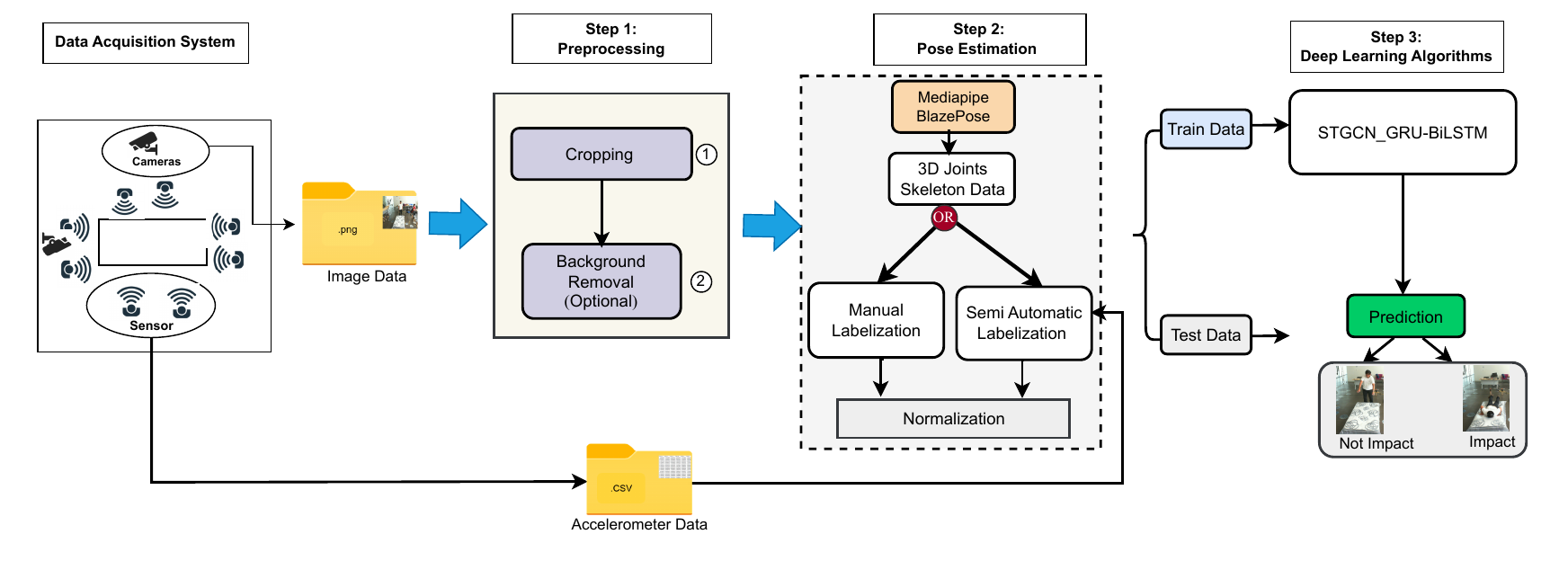}    
\includegraphics[scale=0.31]{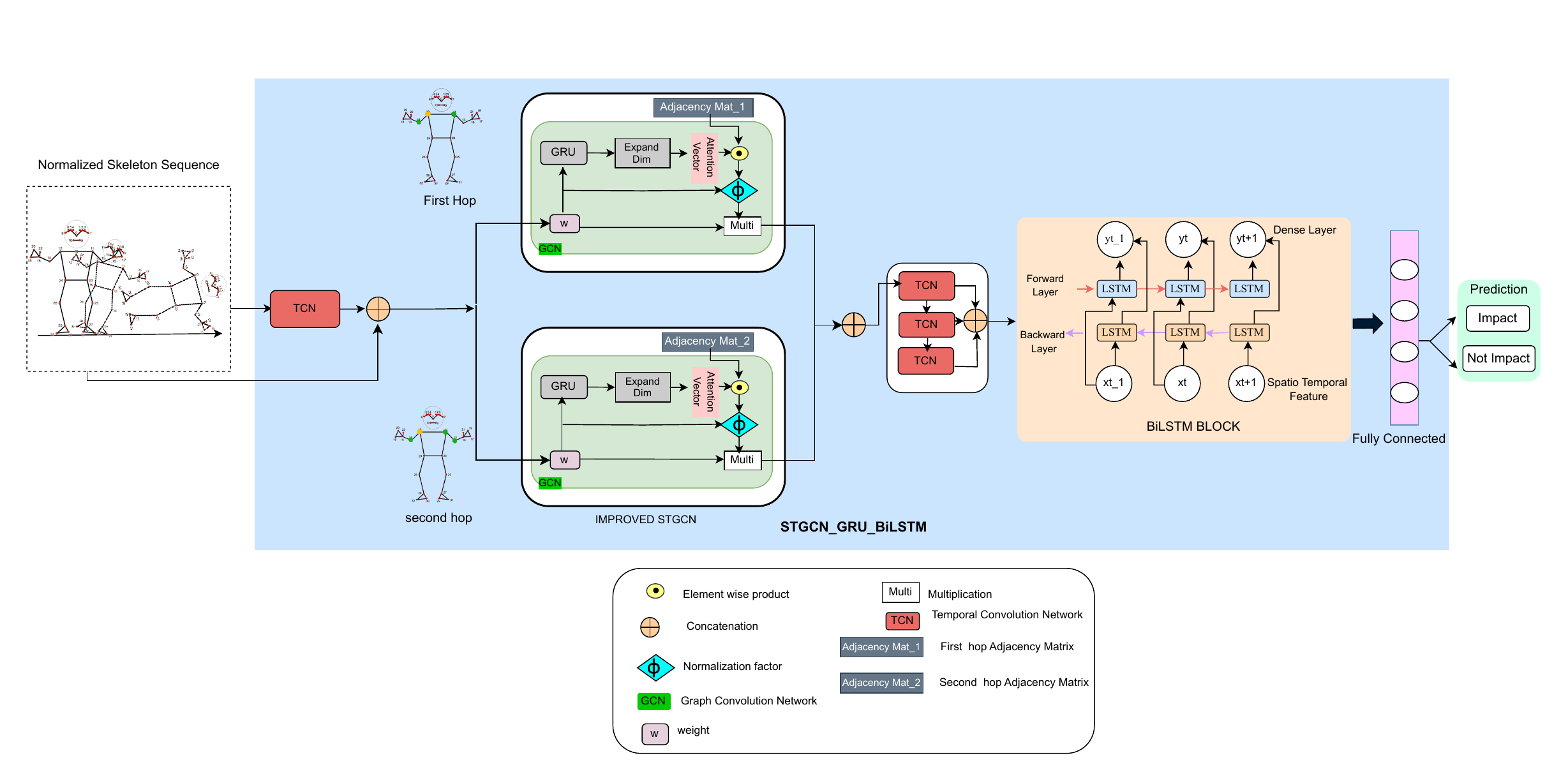}
\captionof{figure}{Flowchart of the proposed approach.}
\label{pipeline}
\end{center}
\subsection{Data Pre-processing}

Data pre-processing is an important step in preparing the input data for the proposed impact detection system. In the context of dealing with multimodal data, including inertial data from accelerometers and synchronized image data, refining the image data is essential to minimize the risk of errors in the analysis. Hence, this step ensures that the image data is processed accurately and aligned appropriately with the inertial data for a comprehensive analysis of fall events.

\subsubsection{Data}
This study utilizes the publicly available UP-Fall dataset, which was originally developed for fall detection research. Although not explicitly designed for impact detection, the dataset offers synchronized accelerometer and image data at a sampling rate sufficient for identifying impact moments with high temporal resolution. It includes a range of fall types, as well as Activities of Daily Living (ADLs) that often resemble fall-like events, thereby enabling the differentiation between impactful and non-impactful occurrences.
Our review of the dataset revealed limitations common to conventional fall detection datasets, including mislabeled events lacking actual ground impact and frames involving multiple individuals, which complicate pose estimation. These issues underscore broader limitations of existing systems, which frequently yield high false positive rates.
Although this dataset was selected for its rich multimodal structure, which can be effectively adapted for impact detection through targeted preprocessing. Specifically, we employ 3D skeleton extraction and semi-automatic relabeling to isolate the primary subject, resolve labeling inconsistencies, and facilitate joint-level motion analysis relevant to ground contact. Our impact-specific framework addresses these limitations by focusing exclusively on events involving verified ground contact.

\subsubsection{Images Cropping}

This step involves carefully isolating the primary subject in the images, removing extraneous background information, and creating focused relevant image segments as shown in Figure \ref{CROP}. Specifically, the cropping process targeted both the left and right sides of the images, creating segments standardized at dimensions of 500 x 700 pixels. This isolation of the main subject, ensures that subsequent analyses and algorithms are applied consistently to a precisely defined area on the targeted subject.
This step is essential for the thorough analysis of the dataset.

\begin{center}
\includegraphics[width=7.5cm]{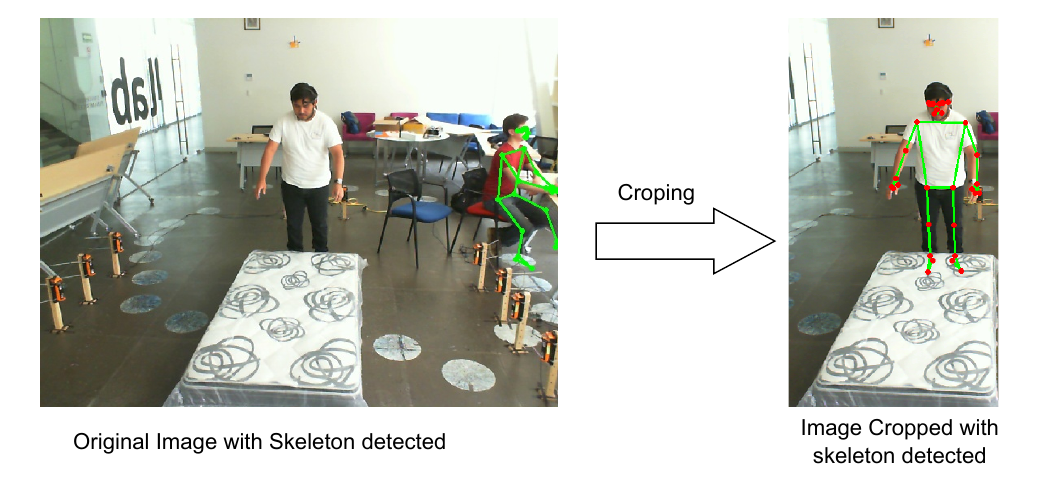}
\captionof{figure}{Image Cropping}
\label{CROP}
\end{center}

\subsubsection{Background Removal}

Background removal is essential for optimizing the number of frames usable for pose estimation. In some cases, pose estimation algorithms might struggle with background elements, leading to dropped frames where 3D joints location can be accurately detected. Therefore, this step is dedicated to eliminating any remaining background elements, ensuring the primary subject is isolated, as shown in Figure \ref{rmved}. Our dataset comprises sequences with varying numbers of frames per activity performed by each subject, with a maximum of 195 frames per sequence.  As shown in Table \ref{tab1}, without background removal, the pose estimation algorithm may detect fewer frames suitable for 3D joints extraction. However, with the application of background removal using the GrabCut algorithm \citep{wang2023review}, the algorithm's performance improve significantly, achieving accurate sleleton detection in up to 100 frames out of the original number of frames. This enhancement substantially increases the number of usable frames and corresponding 3D joints data point available for analysis. This refinement ensures that the pose estimation algorithm can concentrate solely on the subject, thereby enhancing its efficiency and accuracy in detecting frames and extracting 3D joints data points.
\begin{center}
\includegraphics[width=7.5cm]{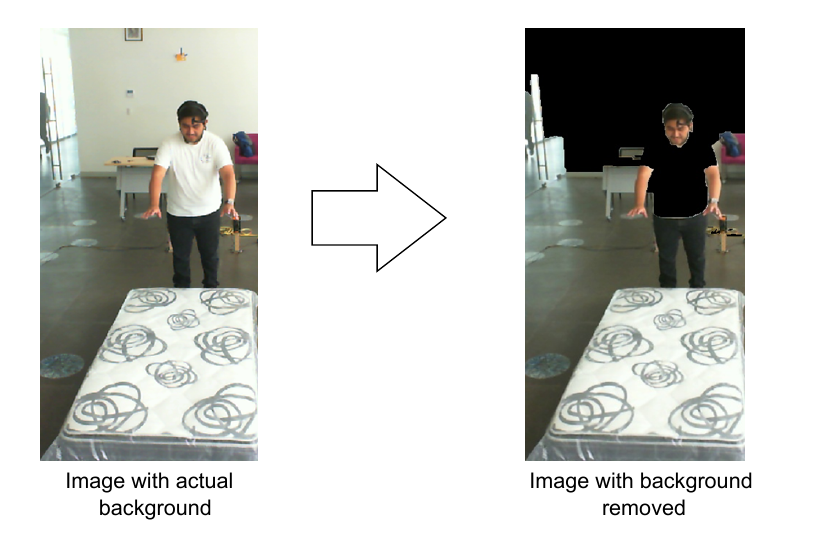}
\captionof{figure}{Background Removal}
\label{rmved}
\end{center}

\begin{center}
\captionof{table}{Impact of Background Removal on Pose Estimation in Detecting Frames on different subjects}
\footnotesize % Adjusts the font size to be smaller than normal but larger than \scriptsize
\label{tab1}
\begin{tabular}{c|c|c|c}
\hline
\textbf{Subject} & \textbf{Total Frames} & \textbf{Frames with Detected Skeleton} & \textbf{Frames with Detected Skeleton} \\
&  & \textbf{(No Background Removal)} & \textbf{(With Background Removal)} \\ \hline
Subject 1 & 195 & 81 & 100 \\ \hline
Subject 2 & 176 & 85 & 110 \\ \hline
Subject 3 & 148 & 78 & 105 \\ \hline
\end{tabular}
\end{center}

\subsection{Extraction of 3D Joint Skeleton Data}

\begin{figure*}[ht]
\centering
\includegraphics[width=12cm]{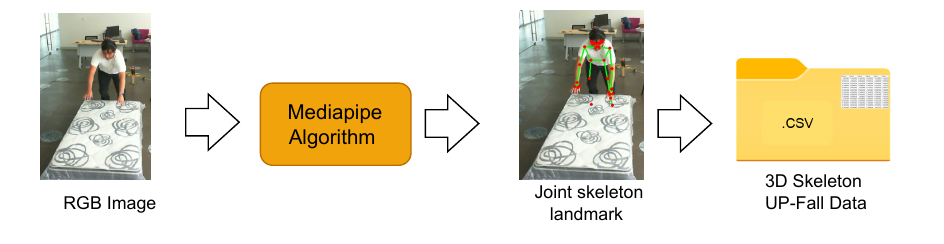}
\caption{Overview of the MediaPipe BlazePose Pipeline for Joint Skeleton Extraction}
\label{mediapipe}
\end{figure*}

The extraction of 3D joint skeletal information plays a critical role in our methodology for detecting impacts within fall events. Understanding the precise positioning and movement of body joints is essential for accurately identifying and analyzing potential impact instances. We evaluated two popular algorithms known for their proficiency in extracting joint information from human subjects: OpenPose~\citep{vyas2019pose} and MediaPipe BlazePose~\citep{grishchenko2022blazepose}.
MediaPipe BlazePose is a state-of-the-art pose estimation model developed by Google's MediaPipe team~\citep{lugaresi2019mediapipe}. It is designed to accurately detect and track human body poses in real-time from images or video streams through its lightweight architecture and optimized inference process, enabling smooth operation across a wide range of devices, including smartphones and embedded systems~\citep{bazarevsky2020blazepose}. The BlazePose algorithm employs a deep neural network architecture trained on large-scale datasets to predict keypoint locations corresponding to various body joints, including shoulders, elbows, wrists, hips, knees, and ankles, totaling 33 joints. These keypoints provide valuable information about spatial configuration and human body movements, enabling applications to analyze gestures, track fitness activities, or detect anomalies such as falls.
OpenPose represents an alternative approach that does not rely on separate framework models~\citep{martinez2019openpose}. It is designed to accurately detect and track human body poses in real-time from images or video streams. OpenPose demonstrates versatility and comprehensiveness in detecting positions of major body joints such as shoulders, elbows, wrists, hips, knees, and ankles, totaling 25 keypoints. For our study focusing on impact detection within fall events, accurate detection of multiple body joints is crucial. We selected MediaPipe BlazePose due to its demonstrated consistency and reliability in capturing joint skeletal data across our dataset. This choice minimizes potential data gaps from algorithmic limitations while BlazePose's proficiency in identifying key anatomical points enables reconstruction of detailed skeletal frameworks for each image. Figure~\ref{mediapipe} illustrates the complete MediaPipe BlazePose pipeline, from RGB image input through joint skeleton landmark detection to the generation of 3D skeleton UP-Fall data. The structured output presents frame-by-frame joint positions in three-dimensional space (X, Y, Z), offering insights into spatial relationships and facilitating multidimensional understanding of human body movements leading to impact events.
Our selection of MediaPipe BlazePose is further supported by its superior performance in pose estimation across our dataset compared to OpenPose, as demonstrated in Table~\ref{tab2}. MediaPipe BlazePose successfully detected poses in 100 frames with 33 joints, compared to OpenPose's 61 frames with 25 joints, providing more comprehensive data for our impact detection analysis.

\begin{table}[!h]
\centering
\caption{Performance Comparison of OpenPose and MediaPipe BlazePose on a Single Subject}
\label{tab2}
\footnotesize
\begin{tabular}{c|c|c}
\hline
\textbf{Pose Algorithm} &  \textbf{Frames Detected}  &  \textbf{Number of Joints}\\
\hline
OpenPose  & 61 & 25 \\ 
\hline
MediaPipe BlazePose  & 100& 33  \\
\hline
\end{tabular}
\end{table}

\subsection{Problem Statement}

In addressing the challenge associated with impact detection within fall events, our focus centers on utilizing 3D skeletons data extracted through the pose estimation algorithm. Each fall event denoted as \(V_i = \{X_t=1...T\}\), represents an arbitrary fall instance, where \(V_i\) is the ith sequence of joint skeleton frames, \(X_t\) represents the t-th frame, and \(T\) is the total number of frames in the fall sequence. Associated with each fall event, a ground-truth label \(y_i\) represents the occurrence of an impact.

Each skeleton comprises \(N\) joints, with each joint having \(C\) dimensional coordinates estimated through the pose estimation algorithm. Consequently, the joint skeleton data is represented as \(V_i \in \mathbb{V}^T \times N \times C\), where \(\mathbb{V}^T\) denotes the set of all possible sequences of joint skeleton frames of length \(T\), and for each frame \(X_t \in \mathbb{R}^{N \times C}\).

Recognizing the significance of individual joint motions during a fall, the primary objective is to accurately detect the moment of impact, denoted \(\hat{y}_j\), signifying the occurrence of contact with the ground.

\subsection{Manual Labelization}

Given that some images in our dataset labeled as ”impact”
did not exhibit indications of an impact upon visual inspection
in the corresponding fall folder as shown in Figure  \ref{semi}, to validate each detected event, we examine the dataset visually, ensuring its integrity and systematic organization. This process involves manually labeling into impact and non-impact data, confirming their categorization and storage within the designated folders. The impact data are labeled as 1, while non-impact data are appropriately labeled as 0. This manual and visual inspection step serves a dual purpose in our data preparation process. On the one hand, it validates the accuracy of the data and ensures proper folder assignment. On the other hand, it contributes to maintaining the entire dataset in good quality and consistency.  However, While manual labeling allows for meticulous visual inspection to determine the presence of impact with high accuracy, it is a time-consuming and labor-intensive process, highlighting the need for more efficient automated methods. 

\begin{center}
\includegraphics[scale=0.55]{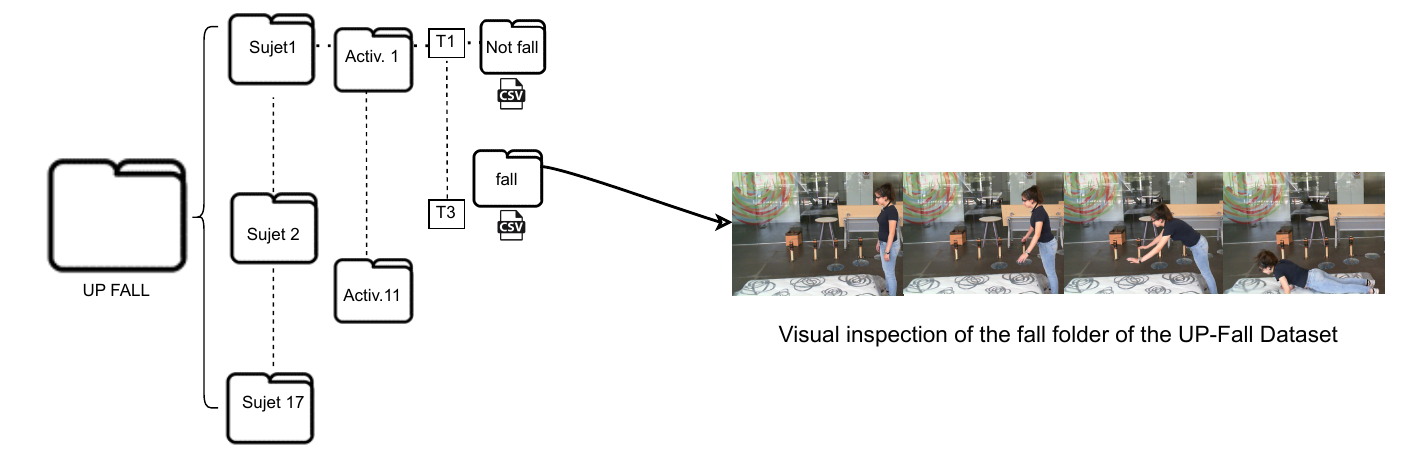}
\captionof{figure}{The Current UP-Fall Dataset for Fall Detection}
\label{semi}
\end{center}

\subsection{Semi-Automatic Labelization}

This section introduces an additional contribution to the dataset aimed to automatically detect impact within fall events using only the inertial data that is captured synchronously with camera recordings, addressing labour intensive and the time-consuming issue from the manual labeling. Given that the dataset is multimodal, consisting of both image data obtained from cameras and inertial data obtained from accelerometers, in this step, we focus on leveraging the inertial data. To achieve this, we initially calculate the Signal Magnitude Vector (SMV) which condenses the information from the accelerometer's three axes (x, y, z) into a single magnitude value using the equation \ref{eq:SMV} \citep{lin2020fall}:

\begin{equation}
\text{SMV} = \sqrt{a_x^2 + a_y^2 + a_z^2} 
\label{eq:SMV}
\end{equation}

Where \( a_x \),\( a_y \) and \( a_z \)represent the acceleration along the x-axis, y-axis and z-axis respectively. To mitigate noise in the data and minimize delays in fall event detection, we compare the calculated SMV value with a predefined threshold to distinguish between impact and non-impact events:

\begin{equation}
\begin{cases}
\text{if SMV} > \beta, \text{ then impact detected} \\
\text{if SMV} \leq \beta, \text{ then no impact detected}
\end{cases}
\end{equation}

Here, $\beta$ represents the given threshold value.

An impact is considered detected if the SMV value exceeds the specified threshold of $2g$, with  $g$ being the gravitational acceleration unit. Hence, to validate each detected event, a visual check is performed by examining the corresponding visual data. This process ensures that the data is accurately labeled, marking impact events as 1 and 0 for non-impact events as shown in Figure \ref{visual}. Furthermore, the validation process guarantees alignment between each detected event and the corresponding frame identified in the visual data.
Upon closer visual inspection, discrepancies between the semi-automatic labeling and the actual impact event can be identified, which can result in false alerts. The proposed semi-automatic approach's robustness lies in its ability to accurately capture the timing of impact events and reduce the number of frames requiring manual inspection. It also helps in managing cross-modal data labeling efficiently by automating the detection of impact events using only inertial data from accelerometers to label frames, ensuring coherence and reliability in the labeled dataset.
\begin{center}
\includegraphics[scale=0.55]{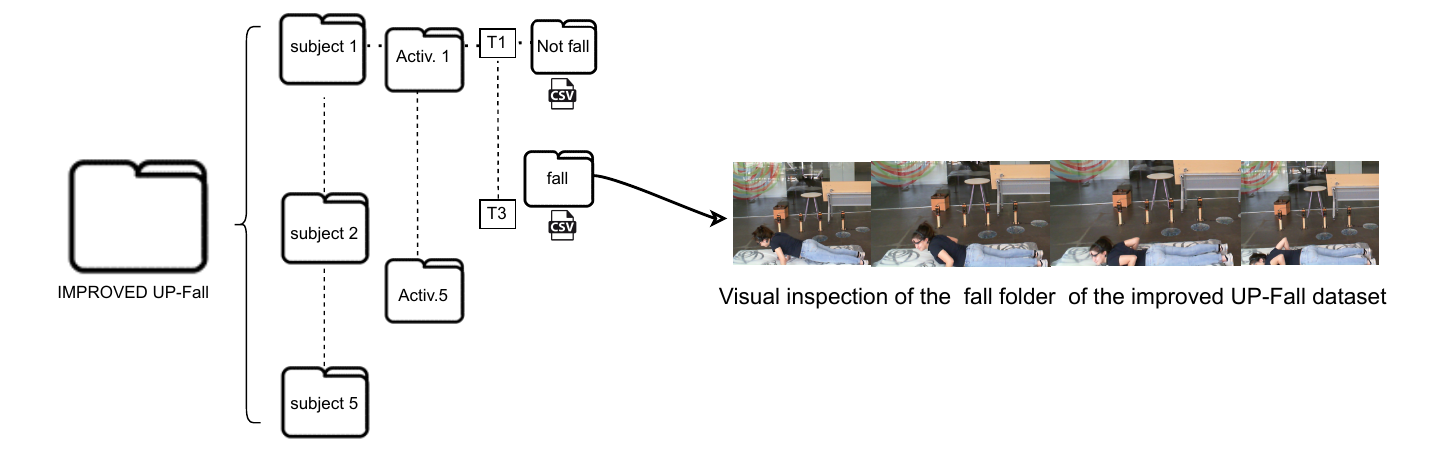}
\captionof{figure}{The Improved UP-FALL Dataset for Impact Fall Detection}
\label{visual}
\end{center}
\subsection{Normalization}

We applied the standard normalization, as defined by \citep{jeong2022fall}, to the joint skeletal data extracted from each image of the fall dataset. Given that the X, Y, and Z coordinates of the joint movements hold different ranges of values, it is essential to apply a normalization technique. The standard normalization formula is given by:
\begin{equation} 
X_{\text{normalized}} = \frac{X - \mu}{\sigma}
\end{equation}

where \(X\) is the original value of the data point, \(\mu\) is the mean of the data, and \(\sigma\) is the standard deviation of the data.
This process ensures that each dimension of movement is brought to a common scale, granting equal weight to all dimensions in the subsequent analyses. The standard normalization contributes to the consistency and reliability of the data, laying a foundation for accurate and unbiased results in the subsequent stages of the impact detection system. This helped in achieving more stable and efficient convergence of the training algorithms.

\subsection{Models Architecture for Impact Detection}

In our study, we examined diverse methodologies for impact detection within fall events, starting with the STGCN baseline model for capturing spatial and temporal dependencies in skeletal data. We then enhanced the architecture by incorporating ConvLSTM, which analyze the movements and positions of body joints, providing a nuanced understanding of human motion. This capability helps differentiate between types of falls based on the dynamics and involvement of different body joints. Further evolution led to the development of the STGCN-GRU model. GRU layers were integrated to provide faster and more efficient analysis of temporal dependencies, significantly reducing computational complexity. These GRU layers also played a crucial role in identifying the contribution of each body joint during fall events, which is essential for differentiating the type and direction of falls.
Finally, we enhanced our model with the STGCN-GRU-BiLSTM architecture, which includes Bidirectional Long Short-Term Memory (BiLSTM) layers. These layers are crucial for accurately pinpointing the timing of impacts in each frame during fall events. By analyzing dependencies from both past and future frames, BiLSTM layers enhance  the model's temporal analysis, allowing it to detect and assess impact events with high precision within dynamic fall scenarios. This ensures that each frame is evaluated in context, providing detailed and reliable impact detection. Hence, initially, we adopted the Spatio-Temporal Graph Convolutional Networks (STGCN) baseline framework\citep{abdo2023human} \citep{keskes2021vision}, known for its ability to capture spatial and temporal dependencies in graph-structured data, particularly skeletal data representing human body movements.

We represent human body-joints at each frame of the fall sequence images as a graph \(G = (A, V, E)\), where \(V\) is the set of body joints, \(E\) is the set of edges (connections between body joints), and \(A \in \mathbb{R}^{N \times N}\) is the adjacency matrix of the graph \(G\). The skeletal data sequence from the body joints is denoted as \(X = \{x_1, x_2, \ldots, x_t\}\), representing the input data, where \(x_t\) refers to the skeletal data at time step \(t\).

This initial processing is written as:
\begin{equation}    
Z = X \oplus (\Gamma^\mu \otimes V) 
\end{equation}
Here, \( Z \in \mathbb{R}^{T \times N \times (C+C_\mu)} \) represents the processed sequence after temporal convolution and concatenation operations. \( C_\mu \) specifies the number of filters in the temporal convolution.
\( \otimes \) and \( \oplus \) 
are respectively the temporal convolution and the concatenation.

To extract spatial features from the skeletal structure, graph convolution is applied to the transformed data \(Z\) using the \(k\)-th hop adjacency , \(A_k\) using the GCN rule update as follows:
\begin{equation}    
G_k = \sigma(\tilde{A}_k Z W_k)
\end{equation}
where \(\tilde{A}_k = D_k^{-\frac{1}{2}} (A_k + I) D_k^{-\frac{1}{2}}\). Here, \(D\) is the diagonal degree matrix, \(W_k\) is the weight matrix and \( \sigma \) is a non-linear activation function that facilitates the linear transformation and aggregation of neighboring features. Following the graph convolution, three Temporal Convolutional (TCNs) using the same padding and kernel size  \( \Gamma^l_1 \), \( \Gamma^l_2 \), and \( \Gamma^l_3 \), respectively to extract different levels of temporal features. To detect impact dynamics across various fall scenarios, we concatenate both higher and lower-level features using the following equations:
\begin{equation}
\begin{split}
    f_1 &= \Gamma^l_1 \otimes G_k, \\ 
    f_2 &= \Gamma^l_2 \otimes f_1, \\ 
    f_3 &= \Gamma^l_3 \otimes f_2, \\ 
    f_k &= f_1 \oplus f_2 \oplus f_3
\end{split}
\end{equation}

where \(f_k\) represents the spatio-temporal features extracted from the k-th hop. The outputs from each hope are then concatenate as: 
\begin{equation}    
Y = f_1 \oplus f_2 \oplus \ldots \oplus f_k
\end{equation}

The output from the Spatio-Temporal Graph Convolutional Network (STGCN) block is represented as \( Y \) with dimensions 
\({R}^{T \times N \times C_{\text{out}}} \), where \( C_{\text{out}} = \sum_{i=1}^{3} kC^l_i \) is  a 3D tensor. To further enhance the model's capability to identify movement patterns and impacts, we employ multiple stacked STGCN blocks to extract more complex features. Finally, global average pooling is applied to the output of the last STGCN block, \(Y_l\), producing the feature vector \(Y_{pool} \in \mathbb{R}^{C_{out}}\). This feature vector is then processed by a series of Fully Connected (FC)layers to predict impact.

The extracted spatio-temporal features, \(Y_{\text{}}\), provide a baseline for impact detection, however, there is still significant room for improvement. Firstly, the use of a global pooling layer before FC layers may ignore the sequential dependencies present among spatio-temporal features across frames. As a result, subjects undergoing the same fall at different speeds may produce distinct spatio-temporal representations. This limitation becomes particularly critical in accurately detecting impacts. Secondly, different types of falls tend to emphasize specific body joints. However, ST-GCN treats all joints equally, which restricts its ability to capture the varying importance of individual joints in impact detection. Therefore, to enhance the temporal modeling capabilities, our first improvement incorporates ConvLSTM layers into the architecture instead of relying solely on the spatial STGCN layers. This variant uses an attention vector denoted as \( \mathbf{Va}_k \in \mathbb{R}^{T \times N} \) to enhance the model's focus on crucial body joints during fall events.
The operation of the attention vector can be mathematically represented as:
\begin{equation}
G_k = \sigma \left( \phi \left( (\hat{A}_k \odot Va_k) Z W_k \right) \right)
\end{equation}
where \(\odot\) denotes the element-wise product, \(\hat{A}_k = A_k + I\) includes the identity matrix to maintain the joint connections, and \(\phi(\cdot)\) acts as a normalizing function to scale the features and stabilize training. 
This attention vector \(Va\) is applied to the adjacency matrix \(A\) by element-wise multiplication which is computed from the neural network outputs concerning current input features. The attention vector \(Va\) is derived through a softmax function that emphasizes the most relevant joints to the impact, resulting in a modified adjacency matrix \( \tilde{A} = Va \odot A \).
This allows for targeted feature extraction that prioritizes critical joint interactions, enhancing the model’s understanding of fall dynamics.  The ConvLSTM layers process the input sequence as follows:
\begin{equation}    
\begin{aligned}
i_t &= \sigma(W_{xi} \otimes X_t + W_{hi} \otimes H_{t-1} + W_{ci} \odot C_{t-1} + b_i) \\
f_t &= \sigma(W_{xf} \otimes X_t + W_{hf} \otimes H_{t-1} + W_{cf} \odot C_{t-1} + b_f) \\
C_t &= f_t \odot C_{t-1} + i_t \odot \tanh(W_{xc} \otimes X_t + W_{hc} \otimes H_{t-1} + b_c) \\
o_t &= \sigma(W_{xo} \otimes X_t + W_{ho} \otimes H_{t-1} + W_{co} \odot C_t + b_o) \\
H_t &= o_t \odot \tanh(C_t)
\end{aligned}
\end{equation}
where \(i_t\) is the input gate, \(f_t\) is the forget gate, \(o_t\) is the output gate, \(C_t\) is the cell state, and \(H_t\) is the hidden state. The symbol \(\otimes\) denotes the convolution operation, \(\odot\) denotes the element-wise multiplication, \(\sigma\) is the sigmoid activation function, \(\tanh\) is the hyperbolic tangent activation function, \(W_{xi}, W_{xf}, W_{xo}, W_{xc}, W_{hi}, W_{hf}, W_{ho}, W_{hc} \in \mathbb{R}^{1 \times 1}\) are the convolutional weight matrices, \(W_{ci}, W_{cf}, W_{co} \in \mathbb{R}^{1 \times 1}\) are the weight matrices for the cell state, and \(b_i, b_f, b_o, b_c\) are the bias parameters.
This enhancement improves the model's ability to analyze joint movements.
As we continually refine our approach for detecting impact within fall events, the second enhancement introduces the combination of Spatio-Temporal Graph Convolutional Networks (STGCN) with Gated Recurrent Unit (GRU) layers, forming the STGCN-GRU architecture. This adaptation addresses the need to reduce the computational complexities while preserving the varying roles of each joint in different types of falls and maintaining the model's effectiveness. In this architecture, GRU layers replace the ConvLSTM due to their more memory-efficient nature while still preserving the crucial ability to extract relevant joint information essential for impact detection within fall events. The GRU mechanism operates as follows:
\begin{equation}
\begin{aligned}
z_t &= \sigma(W_z \cdot [h_{t-1}, x_t]) \\
r_t &= \sigma(W_r \cdot [h_{t-1}, x_t]) \\
\tilde{h}_t &= \tanh(W \cdot [r_t \odot h_{t-1}, x_t]) \\
h_t &= (1 - z_t) \odot h_{t-1} + z_t \odot \tilde{h}_t
\end{aligned}
\end{equation}
where \(h_t\) represents the hidden state at time \(t\). This enables the model to efficiently capture and update temporal information in the skeletal sequence.
Continuing our efforts to improve impact detection in fall events, the third enhancement introduces Bidirectional Long Short-Term Memory (BiLSTM) layers into the STGCN-GRU architecture. This augmentation acknowledges the variability in the time individuals take to reach the impact point, which is an important consideration in fall event scenarios. The integration of BiLSTM layers enhances the model’s temporal understanding by capturing dependencies from both past and future frames. This dual-directional processing enables the model to detect impact events within the dynamic context of fall sequences more effectively. The BiLSTM layers, represented as:
\begin{equation}    
\begin{aligned}
i_t^f &= \sigma(W_{xi}^f x_t + W_{hi}^f h_{t-1}^f + b_i^f) \\
f_t^f &= \sigma(W_{xf}^f x_t + W_{hf}^f h_{t-1}^f + b_f^f) \\
o_t^f &= \sigma(W_{xo}^f x_t + W_{ho}^f h_{t-1}^f + b_o^f) \\
g_t^f &= \tanh(W_{xg}^f x_t + W_{hg}^f h_{t-1}^f + b_g^f) \\
c_t^f &= f_t^f \odot c_{t-1}^f + i_t^f \odot g_t^f \\
h_t^f &= o_t^f \odot \tanh(c_t^f) \\
i_t^b &= \sigma(W_{xi}^b x_t + W_{hi}^b h_{t-1}^b + b_i^b) \\
f_t^b &= \sigma(W_{xf}^b x_t + W_{hf}^b h_{t-1}^b + b_f^b) \\
o_t^b &= \sigma(W_{xo}^b x_t + W_{ho}^b h_{t-1}^b + b_o^b) \\
g_t^b &= \tanh(W_{xg}^b x_t + W_{hg}^b h_{t-1}^b + b_g^b) \\
c_t^b &= f_t^b \odot c_{t-1}^b + i_t^b \odot g_t^b \\
h_t^b &= o_t^b \odot \tanh(c_t^b)
\end{aligned}
\end{equation}
where \(h_t^f, c_t^f\) and \(h_t^b, c_t^b\) denote the hidden and cell states for forward and backward directions, respectively, facilitate the model’s ability to capture long-term dependencies and context information from both past and future frames. By combining forward and backward states, the STGCN-GRU-BiLSTM architecture offers a comprehensive solution tailored to the complexities of fall event impact detection. The integration of BiLSTM layers into the STGCN-GRU architecture represents a significant step towards refining the model’s temporal sensitivity and addressing the complexities inherent in fall event scenarios. 
Thus, this architecture, as shown in Figure \ref{pipeline}, appears to be the improvement of the state-of-the-art fall detection system dealing with detecting impact within fall events, offering a practical solution for real-world applications.  The commonly used symbols and notations in the equations and figures of our paper are summarized in Table \ref{tab3}.

\begin{center}
\captionof{table}{Summary of Common Notations Used}
\label{tab3}
% \footnotesize % Adjusts the font size to be smaller than normal but larger than \scriptsize
\begin{tabular}{cl}
\hline
\textbf{Notation} & \textbf{Definition} \\ \hline
$\odot$ & Element-wise product \\
$\otimes$ & Temporal convolution operation \\
$\oplus$ & Concatenation operation \\
$G = (A, V, E)$ & Graph representation of body joints \\
$A \in \mathbb{R}^{N \times N}$ & Adjacency matrix of the graph $G$ \\
$V$ & Set of vertices (body joints) \\
$E$ & Set of edges \\
$C_\mu$ & Number of filters in temporal convolution \\
$G_k$ & Output of graph convolution at hop $k$ \\
$\sigma$ & Nonlinear activation function \\
$W_k$ & Learnable weight matrix for hop $k$ \\
$\tilde{A}_k$ & Normalized adjacency matrix for hop $k$ \\
$D_k$ & Diagonal degree matrix for hop $k$ \\
$f_k$ & Spatio-temporal features\\
$Y \in \mathbb{R}^{T \times N \times C_{out}}$ & Final output of the STGCN block \\
$C_{out}$ & Dimension of output channels in the final tensor \\
$Y_{pool} \in \mathbb{R}^{C_{out}}$ & Feature vector after global average pooling \\
$H_t, C_t$ & Hidden and cell states at time $t$\\
$h_t$ & Hidden state at time $t$ in GRU \\
$h^f_t, c^f_t, h^b_t, c^b_t$ & Forward and backward hidden layers in BiLSTM \\
$i_t, f_t, o_t, g_t$ & Input, forget, output, and gate activation vectors \\
$W_{xi}, W_{xf}, W_{xo}, W_{xc}$ & Weight input, forget, output, and cell gates \\
$b_i, b_f, b_o, b_c$ & Bias parameters for input, forget, output \\
\hline
\end{tabular}
\end{center}

\section{Experimentation and Results}
\label{sec:Experimental Results}

In this section, we delve into the experimentation and results of our proposed impact detection model. We begin by introducing the dataset and metrics used for evaluating the effectiveness of our approach. The skeletal data obtained from the Mediapipe BlazePose algorithm, as mentioned earlier, undergoes labeling through either a manual or semi-automatic process based on our proposed pipeline in Figure \ref{pipeline}. This labeling process sets the stage for our experiments, allowing us to assess the model's performance and its effectiveness.

\subsection{Data Description}

Obtaining real-world data from the elderly is still challenging due to privacy issues. To address this limitation, our study utilizes two  datasets publicly available, the UP-Fall \citep{martinez2019up} and UMAFall datasets \citep{casilari2017umafall}. These datasets serve as the foundation for our research, providing a diverse range of simulated fall scenarios. We then apply our  preprocessing method to extract 3D skeleton data, enhancing the utility of these datasets for our specific research objectives.

\textbf{UP-Fall dataset} comprises falls and daily activities data from 17 healthy young adults (9 males, 8 females, ages 18-24). Collected using a multimodal approach with 14 devices, it incorporates wearable sensors (accelerometers, gyroscopes, ambient light sensors) and cameras capturing high-resolution lateral and frontal views \citep{martinez2019up}. As shown in Table \ref{tab4}, the dataset includes 11 activities: daily activities and various fall scenarios. The dataset focuses on body point alterations during falls, predominantly featuring lateral camera images of forward and sideward falls. Images were recorded at approximately 18 Hz, providing a detailed time-sequenced dataset.
\begin{center}
\captionof{table}{Activities performed by subjects, adapted in UP-Fall Dataset\citep{martinez2019up}}
\label{tab4}
\scriptsize
\resizebox{0.8\textwidth}{!}{%
\begin{tabular}{cc p{5cm} cc}
\hline
\textbf{Activity} & \textbf{Category} & \textbf{Description} & \textbf{Duration (s)} & \textbf{Abbreviation} \\
\hline
1 & FALL & Falling forward using hands (\textbf{Fall forward})  & 10 & FH \\
2 & FALL & Falling forward using knees & 10 & FF \\
3 & FALL & Falling backward (\textbf{Fall backward}) & 10 & FB \\
4 & FALL & Falling sideward (\textbf{Fall Lateral}) & 10 & FS \\
5 & FALL & Falling while attempting to sit in an empty chair & 10 & FE \\
\hline
6 & ADL & Walking & 60 & W \\
7 & ADL & Standing & 60 & S \\
8 & ADL & Sitting & 60 & ST \\
9 & ADL & Picking up an object & 10 & P \\
10 & ADL & Jumping & 30 & J \\
11 & ADL & Laying & 60 & L \\
\hline
\end{tabular}%
}
\end{center}

\begin{center}
\captionof{table}{Activities Performed by subjects in UMAFall Dataset\citep{casilari2017umafall}}
\label{tab5}
\footnotesize
\begin{tabular}{ccc}
\hline
\textbf{Activity ID} & \textbf{Category} & \textbf{Description} \\
\hline
1 & FALL & Falling forward  \\
2 & FALL & Falling Backward \\
3 & FALL & Falling lateral \\ \hline
4 & ADL & climbing stairs \\
5 & ADL & raising the hands \\
6 & ADL & Walking \\
7 & ADL & Standing \\
8 & ADL & Sitting \\
9 & ADL & jogging \\
10 & ADL & body bending \\
11 & ADL & lying down and getting up from a bed \\
12 & ADL & clapping hands \\
13 & ADL & making a phone call \\
14 & ADL & opening a door \\
15 & ADL & Jumping \\
16 & ADL & Laying \\
\hline
\end{tabular}
\end{center}

\textbf{UMAFall dataset} is another publicly available multimodal dataset curated by Casilari et al. \citep{casilari2017umafall}. It is acquired through systematic emulation of a range of predefined Activities of Daily Life (ADLs) and falls. In contrast to the UP-Fall dataset, which includes five types of falls, the UMAFall dataset consists of three distinct types of falls such as lateral, frontal, and backward falls as described in Table \ref{tab5}. These three types of falls in the UMAFall dataset correspond to similar types of falls in the UP-Fall dataset, such as "FB+ Standing," "Fside+standing," and "FH+ Standing." This similarity allows for direct comparison and evaluation of fall detection algorithms across different datasets, providing valuable insights into the robustness and generalizability of the model.

\begin{center}
\small
\captionof{table}{Model Training Parameters}
\label{tab6}
\begin{tabular}{l l}
\hline
\textbf{Network Parameters} & \textbf{Values} \\
\hline
Epochs & 300 \\
Batch size & 8 \\
Dropout & 0.25 \\
Learning rate & 0.0001 \\
Activation Function & Sigmoid \\
Optimization Algorithm & Adam \\
Loss Function & Binary Cross Entropy \\
k-hops & 2 \\
Hidden size  & 80, 40, 40, 80 \\
GRU units & 33, 33 \\
Input joints & 33 \\
Input channels & 3 \\
Sequence length & 100 \\
\hline
\end{tabular}
\end{center}
\subsection{Training and Testing Strategy}

For our evaluation protocol, we divided the 3D skeletons from the UP-Fall dataset into three distinct subsets: a training set (80\%), a validation set (10\%), and a test set (10\%). The training set was used to train the model, while the validation and test sets were used to assess performance and generalization. To ensure a diverse and representative evaluation, we selected two fall sequences from each subject for inclusion in the test set. This strategy helps evaluate the model's ability to generalize across different individuals and fall scenarios.
For temporal sequence processing, we employed a fixed window size of 100 frames per fall sequence, corresponding to approximately 5.6 seconds at the UP-Fall dataset's 18 Hz sampling rate. This window size was determined through comprehensive analysis of our pose estimation capabilities across the entire dataset. While the original UP-Fall sequences contain varying numbers of frames ranging from activity-dependent durations, our MediaPipe BlazePose algorithm consistently detected reliable 3D skeleton poses in at least 100 frames per sequence across all subjects and activities, regardless of the original sequence length. To ensure temporal consistency and maximize data utilization across all sequences, we standardized the window size to 100 frames, representing the guaranteed minimum number of frames with accurately detected joint positions available for every sequence in the dataset.
During the parameter setup detailed in Table~\ref{tab6}, we conducted extensive experiments and ultimately selected a two-hop adjacency matrix (k = 2). This configuration enables the model to capture interactions not only between immediate neighboring joints but also among secondary neighbors, enhancing its capacity to model complex joint dependencies.

\subsection{Evaluation Metrics}

For the experiments, we measure the performance of the classification models using eight metrics such as accuracy, precision, sensitivity, specificity, F1-score, recall, ROC, MCC \citep{vujovic2021classification}, as shown in Equations (12)–(19), where $TP$ and $TN$ are respectively the true positives and true negatives while $FP$ and $FN$ are the false positives and false negatives respectively.

\begin{equation}
\text{accuracy} = \frac{\text{TP} + \text{TN}}{\text{TP} + \text{TN} + \text{FP} + \text{FN}} \label{eq:accuracy}
\end{equation}

Equation \ref{eq:accuracy} calculates the accuracy of the classification model in correctly identifying both impact and non-impact events within fall data. It assesses the model's overall correctness by considering the number of true positive (correctly identified impacts) and true negative (correctly identified non-impacts) predictions relative to all instances, including false positives (non-impacts incorrectly classified as impacts) and false negatives (impacts incorrectly classified as non-impacts). Higher accuracy values indicate a higher proportion of correctly classified impact and non-impact events, reflecting the model's effectiveness in distinguishing between the two categories.

\begin{equation}
\text{precision} = \frac{TP}{TP + FP} \label{eq:precision}
\end{equation}

Equation \ref{eq:precision} computes precision, this metric assessing the accuracy of positive predictions in distinguishing impact events from non-impact events within fall data. It measures how well the model performs when predicting an impact event (positive class). Higher precision values indicate fewer false positives, demonstrating the model's ability to accurately identify true positive impact events while minimizing incorrect predictions of non-impact events.

\begin{equation}
\text{sensitivity} = \frac{TP}{TP + FN} \label{eq:sensitivity}
\end{equation}

Equation \ref{eq:sensitivity} computes sensitivity, a critical metric in impact detection within fall events. It quantifies the model's ability to accurately identify positive impacts among all actual positive instances. Higher sensitivity values indicate a better ability to detect true positive impacts, which is essential for ensuring the safety and well-being of individuals experiencing falls.

\begin{equation}
\text{specificity} = \frac{TN}{TN + FP} \label{eq:specificity}
\end{equation}

Equation \ref{eq:specificity} calculates specificity, a crucial metric in impact detection within fall events. It measures the model's ability to correctly identify negative impacts by accurately recognizing true negatives among all actual negative instances. A higher specificity indicates a better ability to distinguish non-impact instances, contributing to the overall performance of the impact detection system.

\begin{equation}
\text{F1-score} = \frac{2 \cdot \text{precision} \cdot \text{sensitivity}}{\text{precision} + \text{sensitivity}} \label{eq:f1-score}
\end{equation}

Equation \ref{eq:f1-score} calculates the F1-score, a metric specifically designed for impact detection within fall events. It combines both precision and sensitivity to provide a balanced assessment of the model's performance in identifying impacts accurately. A higher F1-score indicates better overall performance in detecting impacts while considering both false positives and false negatives.

\begin{equation}
\text{recall} = \frac{TP}{TP + FN} \label{eq:recall}
\end{equation}

Equation \ref{eq:recall}  recall and sensitivity are mathematically equivalent. They measures the model's ability to correctly identify positive class (impact) among all actual positive instances.

\begin{equation}
\text{AUC-ROC} = \text{Area Under the ROC curve} \label{eq:auc-roc}
\end{equation}

Equation \ref{eq:auc-roc} represents the area under the Receiver Operating Characteristics (ROC) curve, specifically tailored for impact detection within fall events. It serves as a metric to quantify the model's ability to distinguish between positive (impact) and negative instances in fall detection. A higher AUC-ROC value indicates that the model performs better at correctly identifying impacts from non-impact events.

\begin{equation}
\text{MCC} = \frac{TP \cdot TN - FP \cdot FN}{\sqrt{(TP + FP)(TP + FN)(TN + FP)(TN + FN)}} \label{eq:mcc}
\end{equation}

Equation \ref{eq:mcc} calculates the Matthews Correlation Coefficient (MCC) \citep{chicco2020advantages}, a metric for evaluating the performance of a classification model in impact detection. 
In the context of impact detection, the MCC serves as a comprehensive measure, considering both positive and negative predictions along with their corresponding true values. The MCC is particularly valuable for assessing the overall accuracy and reliability of the impact detection model.

\begin{table}[!htbp]
\centering
\caption{Ablation Study: Component Contribution Analysis for Impact Detection. The bold values refer to the best overall performance.}
\label{tab7}
\footnotesize
\begin{tabular}{lcccc}
\hline
\textbf{Model Configuration} & \textbf{Components} & \textbf{Accuracy} & \textbf{Precision} & \textbf{F1-Score} \\
& & \textbf{(\%)} & \textbf{(\%)} & \textbf{(\%)} \\
\hline
Baseline & STGCN only & 92.16 & 94.16 & 91.31 \\
+ConvLSTM & STGCN + ConvLSTM & 93.00 & 95.00 & 92.14 \\
+GRU (replace ConvLSTM) & STGCN + GRU & 93.50 & 93.36 & 92.68 \\
\textbf{Full Model (Ours)} & \textbf{STGCN+GRU+BiLSTM} & \textbf{97.50} & \textbf{98.11} & \textbf{97.26} \\
\hline
\multicolumn{2}{l}{\textbf{Improvement over baseline:}} & \textbf{+5.34} & \textbf{+3.95} & \textbf{+5.95} \\
\hline
\end{tabular}
\end{table}

\subsection{Ablation Study}
In our ablation study, as summarized in Table \ref{tab7}, we conducted a systematic investigation to evaluate the contribution of key architectural components within our impact detection model. The aim was to discern the influence of each component on the model's capacity to accurately identify impact instances during falls. We evaluated four different model variants, each representing a unique configuration of the STGCN-based architecture tailored for impact detection.
We first established the importance of using STGCN with average pooling structure. This baseline variant achieved an accuracy of 92.16\%, precision of 94.16\%, and F1-Score of 91.31\%, providing fundamental spatial-temporal graph convolution capabilities without additional enhancements.
Next, we investigated the role of ConvLSTM layers, which capture both spatial and temporal dependencies concurrently. This variant showed improved accuracy of 93.00\%, with the highest precision of 95.00\% and F1-Score of 92.14\%. This enhancement demonstrates that ConvLSTM layers excel in analyzing joint movements and positions over varying time frames, providing detailed examination of human motion during fall events.
We then made a strategic decision to replace ConvLSTM layers with GRU layers, leveraging GRU's ability to effectively model temporal dynamics while maintaining computational efficiency. This transition achieved accuracy of 93.50\%, precision of 93.36\%, and F1-Score of 92.68\%. The GRU layers excel in capturing temporal dynamics of joint movements, allowing the model to identify and analyze the contribution of each body joint during fall events.
Finally, we combined GRU and BiLSTM layers in the STGCN-GRU-BiLSTM variant to achieve comprehensive analysis of fall sequences. This final variant achieved the highest overall performance with accuracy of 97.50\%, precision of 98.11\%, and F1-Score of 97.26\%. The BiLSTM component provides bidirectional temporal analysis, capturing detailed temporal dynamics crucial for accurate impact detection.
The progressive improvement from 92.16\% to 97.50\% accuracy demonstrates how strategic architectural enhancements significantly improve performance in detecting impacts within fall scenarios, with the final model showing a substantial 5.34\% improvement over the baseline.

\subsection{Impact Detection Performance with State-of-the-Art Methods Using the Improved 3D Skeleton UP-Fall Dataset}

This section evaluates the effectiveness of our proposed STGCN-GRU-BiLSTM model against state-of-the-art algorithms on individual fall scenarios. All models were trained and tested using our improved UP-Fall dataset, preprocessed with our semi-automatic labeling technique. The evaluation comprised five distinct fall types tested separately: falling backward while standing (FB+Standing), falling forward using knees (FKnee+Standing), falling forward using hands (FH+Standing), falling while sitting (F+Sitting), and falling sideward (Fside+Standing).
Table~\ref{tab8} presents the comprehensive performance comparison across all fall scenarios. Our proposed method demonstrates superior performance across various fall scenarios, achieving the highest accuracy for falls backward while standing (96.50\%) and falls while sitting (97.50\%). 

These results can be attributed to the distinctive body movement patterns associated with these fall types, which are effectively captured by our multi-component architecture.
For falls backward while standing, the spatial configurations are efficiently processed by the Spatio-Temporal Graph Convolutional Network (STGCN) component, while the Gated Recurrent Unit (GRU) layers capture the temporal dynamics of joint movements. For falls while sitting, where movements are more subtle, the Bidirectional Long Short-Term Memory (BiLSTM) layer provides critical bidirectional analysis, capturing both past and future temporal context. This enables the model to distinguish effectively between normal sitting motions and actual impact events, contributing to enhanced detection accuracy for seated falls.

\begin{center}
\captionof{table}{Individual Fall Scenario Performance Analysis Using State-of-the-Art Algorithms on Improved UP-Fall Dataset. (Abbreviations in the table have the following meaning: FB - Falling Backward, FH - Falling Forward using Hands, FKnee - Falling
Forward using Knees, FSide - Falling Sideward, FSitting - Falling while Attempting to Sit in an Empty Chair). Bold values indicate highest performance.}
\label{tab8}
\resizebox{\textwidth}{!}{%
\begin{tabular}{llccccccc}
\toprule
\textbf{Activity} & \textbf{Algorithms} & \textbf{Accuracy} & \textbf{Precision} & \textbf{F1 Score} & \textbf{Specificity} & \textbf{Recall} & \textbf{AUC ROC} & \textbf{MCC} \\
& & (\%) & (\%) & (\%) & (\%) & (\%) & (\%) & (\%) \\
\midrule
\multirow{5}{*}{\textbf{FB+ Standing}} 
& CNN \citep{ha2024fall} & 94.50 & \textbf{96.07} & 93.76 & 84.51 & \textbf{100} & 96.17 & 88.24 \\
& LSTM \citep{castro2023fall} & 92.50 & 87.50 & 90.31 & \textbf{100} & 90.32 & 95.16 & 82.31 \\
& BiLSTM \citep{csengul2022deep} & 94.00 & 93.93 & 93.22 & 88.24 & 94.97 & 92.60 & 86.53 \\
& STGCN \citep{abdo2023human} & 90.00 & 88.90 & 89.58 & 95.69 & 86.61 & 91.25 & 80.12 \\
& Ours & \textbf{96.50} & 96.05 & \textbf{96.17} & 95.77 & 95.77 & \textbf{99.09} & \textbf{92.39} \\
\midrule
\multirow{5}{*}{FKnee+Standing} 
& CNN \citep{ha2024fall} & 84.50 & 75.78 & 78.91 & 84.50 & 83.23 & 90.72 & 64.80 \\
& LSTM \citep{castro2023fall} & 85.50 & 75.78 & 78.91 & \textbf{100} & 81.44 & 90.72 & 64.80 \\
& BiLSTM \citep{csengul2022deep} & 82.50 & 74.26 & 76.82 & \textbf{100} & 79.04 & 89.52 & 61.93 \\
& STGCN \citep{abdo2023human} & 80.00 & 72.60 & 74.33 & \textbf{100} & 88.02 & 97.64 & 66.59 \\
& Ours & \textbf{86.00} & \textbf{77.05} & \textbf{80.53} & \textbf{100} & \textbf{89.76} & \textbf{97.66} & \textbf{67.10} \\
\midrule
\multirow{5}{*}{FH+Standing} 
& CNN \citep{ha2024fall} & 87.00 & 90.71 & 85.78 & 69.77 & \textbf{100} & 95.41 & 75.37 \\
& LSTM \citep{castro2023fall} & 89.50 & 92.22 & 88.83 & 75.58 & \textbf{100} & 87.79 & 79.89 \\
& BiLSTM \citep{csengul2022deep} & 90.00 & 92.86 & 89.01 & 75.00 & 88.37 & 88.50 & 80.18 \\
& STGCN \citep{abdo2023human} & 87.00 & 90.71 & 85.98 & 69.77 & 98.15 & 84.88 & 75.37 \\
& Ours & \textbf{90.00} & \textbf{93.65} & \textbf{89.49} & \textbf{79.07} & 98.25 & \textbf{88.66} & \textbf{80.25} \\
\midrule
\multirow{5}{*}{\textbf{F+ Sitting}} 
& CNN \citep{ha2024fall} & 94.00 & 95.68 & 93.27 & 83.27 & \textbf{100} & \textbf{99.96} & 87.38 \\
& LSTM \citep{castro2023fall} & 76.50 & 74.88 & 75.25 & 73.97 & 77.95 & 75.96 & 50.83 \\
& BiLSTM \citep{csengul2022deep} & 84.50 & 75.78 & 78.91 & 84.50 & 89.76 & 90.72 & 64.80 \\
& STGCN \citep{abdo2023human} & 92.16 & 94.16 & 91.31 & 81.08 & 95.16 & 90.54 & 82.48 \\
& Ours & \textbf{97.50} & \textbf{98.11} & \textbf{97.26} & \textbf{93.15} & \textbf{100} & 96.58 & \textbf{94.67} \\
\midrule
\multirow{5}{*}{Fside+standing} 
& CNN \citep{ha2024fall} & 88.00 & 79.31 & 83.06 & \textbf{100} & 85.54 & 92.77 & 70.81 \\
& LSTM \citep{castro2023fall} & 82.00 & 74.60 & 76.90 & 92.11 & 79.63 & 85.87 & 59.41 \\
& BiLSTM \citep{csengul2022deep} & 89.50 & 82.20 & 85.71 & \textbf{100} & 87.04 & \textbf{93.83} & 74.87 \\
& STGCN \citep{abdo2023human} & 87.50 & 78.33 & 81.92 & \textbf{100} & 84.34 & 92.17 & 69.13 \\
& Ours & \textbf{90.00} & \textbf{82.76} & \textbf{86.29} & \textbf{100} & \textbf{87.65} & \textbf{93.83} & \textbf{75.78} \\
\bottomrule
\end{tabular}%
}
\end{center}

\subsection{\centering State-of-the-Art Algorithm Performance: Generalization and Preprocessing Analysis}

Recently, several algorithms such as Convolutional Neural Networks (CNN)~\cite{ha2024fall,eltahir2023deep}, Long Short-Term Memory (LSTM)~\cite{castro2023fall,lau2022fall}, and Bidirectional Long Short-Term Memory (BiLSTM)~\cite{csengul2022deep} have been utilized in fall detection tasks. While these approaches have demonstrated effectiveness in general fall detection tasks, their performance in precise impact detection the critical moment when a person contacts the ground remains underexplored. 

% This section conducts a comparative analysis between these state-of-the-art algorithms and our proposed STGCN-GRU-BiLSTM method through two complementary experimental frameworks designed to illuminate both generalization capabilities and preprocessing effectiveness.

This section evaluates these state-of-the-art algorithms alongside our proposed method in impact detection using two distinct experimental scenarios.

The first experimental scenario examines cross-dataset generalization performance. This evaluation protocol involves training all algorithms exclusively on our improved 3D skeleton UP-Fall dataset and testing them on individual fall scenarios from the UMAFall dataset. This evaluation framework is significant as it simulates real-world deployment conditions where models encounter previously unseen from a completely different dataset, including lateral, forward, and backward falls that correspond to similar types in the UP-Fall dataset. 
By using the UMAFall dataset exclusively for testing purposes, this approach provides an unbiased assessment of how well each algorithm can transfer learned impact detection patterns to completely novel scenarios.

The second experimental evaluate the direct effect of our preprocessing methodology on all algorithms by comparing their performance on the original UP-Fall dataset versus our improved UP-Fall dataset. All algorithms were trained and tested on the original UP-Fall dataset (without preprocessing) and separately trained and tested on our improved UP-Fall dataset (with preprocessing). This experimental isolates the specific contributions of our data enhancement techniques, including advanced background removal using the GrabCut algorithm, optimized image cropping procedures, and our novel semi-automatic labeling approach based on accelerometer Signal Magnitude Vector analysis. The comparison encompasses all tested algorithms to demonstrate whether our preprocessing improvements provide universal benefits or exhibit algorithm-specific advantages.

The comprehensive results presented in Table~\ref{tab9} and Table~\ref{tab10} reveal compelling insights into both generalization capabilities and preprocessing effectiveness. In the cross-dataset evaluation, our proposed STGCN-GRU-BiLSTM method demonstrates exceptional robustness, achieving consistently high accuracy rates of 95-97\% across all tested fall scenarios from the UMAFall dataset. This performance significantly surpasses traditional LSTM approaches. The superior performance of our method can be attributed to the synergistic combination of spatial-temporal graph convolution for capturing complex joint relationships, GRU layers for efficient temporal modeling, and BiLSTM components for comprehensive bidirectional sequence analysis.
The preprocessing impact analysis reveals transformative improvements across all evaluated algorithms, providing compelling evidence for the universal applicability of our data enhancement approach. Most notably, the baseline STGCN architecture experiences the most dramatic performance enhancement, with accuracy improving from 72\% on the original dataset to 87.43\% on our improved dataset. This improvement demonstrates that our preprocessing methodology addresses fundamental data quality issues that particularly affect graph-based approaches. Meanwhile, our proposed method maintains consistently superior performance, advancing from 95\% to 97\% accuracy, indicating that our architecture design and preprocessing approach work synergistically to achieve optimal impact detection performance.
These results validate both the effectiveness of our data preprocessing approach and the superiority of our proposed architecture for impact detection. The consistent performance gains observed across all state-of-the-art algorithms when using our improved dataset demonstrate the broad applicability and value of our preprocessing methodology for impact fall detection systems.
% These results validate both the effectiveness of our data preprocessing approach and the superiority of our proposed architecture for impact detection. The consistent performance gains observed across all state-of-the-art algorithms when using our improved dataset demonstrate the broad applicability and value of our preprocessing methodology for impact fall detection systems.
% These results validate both the effectiveness of our data preprocessing approach and the superiority of our proposed architecture for impact detection. The consistent performance gains observed across all state-of-the-art algorithms when using our improved dataset demonstrate the broad applicability and value of our preprocessing methodology for impact fall detection systems.

\begin{center}
\captionof{table}{Generalization Performance Evaluation of State-of-the-Art Algorithms: Training on Improved UP-Fall, Testing on UMAFall Individual Scenarios. Bold values indicate highest performance in each metric.}
\label{tab9}
\resizebox{\textwidth}{!}{%
\begin{tabular}{llccccccc}
\toprule
\textbf{Activity} & \textbf{Algorithms} & \textbf{Accuracy} & \textbf{Precision} & \textbf{F1 Score} & \textbf{Specificity} & \textbf{Recall} & \textbf{AUC ROC} & \textbf{MCC} \\
& & (\%) & (\%) & (\%) & (\%) & (\%) & (\%) & (\%) \\
\midrule
\multirow{5}{*}{FB+ Standing} 
& CNN \citep{ha2024fall} & 90.00 & 92.86 & 89.01 & 75.00 & \textbf{100} & 87.50 & 80.18 \\
& LSTM \citep{castro2023fall} & 72.50 & 71.42 & 71.53 & 67.50 & 75.83 & 71.67 & 43.08 \\
& BiLSTM \citep{csengul2022deep} & 92.50 & 94.44 & 91.89 & 81.25 & \textbf{100} & 90.62 & 84.98 \\
& STGCN \citep{abdo2023human} & 90.00 & 92.54 & 89.39 & 76.74 & 88.37 & 88.37 & 79.81 \\
& \textbf{Ours} & \textbf{97.00} & \textbf{97.86} & \textbf{96.52} & \textbf{90.91} & \textbf{100} & \textbf{95.45} & \textbf{93.28} \\
\midrule
\multirow{5}{*}{FH+Standing} 
& CNN \citep{ha2024fall} & 90.00 & 92.86 & 89.01 & 75.00 & \textbf{100} & 87.50 & 80.18 \\
& LSTM \citep{castro2023fall} & 89.50 & 88.21 & 87.67 & 80.95 & 93.43 & 87.19 & 75.40 \\
& BiLSTM \citep{csengul2022deep} & 91.00 & 91.46 & 89.07 & 77.78 & 97.08 & 87.43 & 78.79 \\
& STGCN \citep{abdo2023human} & 94.00 & 95.77 & 93.11 & 82.86 & \textbf{100} & 91.43 & 87.09 \\
& \textbf{Ours} & \textbf{95.00} & \textbf{96.43} & \textbf{94.30} & \textbf{85.71} & \textbf{100} & \textbf{92.86} & \textbf{89.21} \\
\midrule
\multirow{5}{*}{Fside+standing} 
& CNN \citep{ha2024fall} & 90.00 & 88.67 & 88.31 & 82.54 & 93.43 & 87.99 & 76.65 \\
& LSTM \citep{castro2023fall} & 93.50 & 95.36 & 92.68 & 82.19 & \textbf{100} & 91.10 & 86.35 \\
& BiLSTM \citep{csengul2022deep} & 93.50 & 95.58 & 92.22 & 80.30 & \textbf{100} & 90.15 & 85.56 \\
& STGCN \citep{abdo2023human} & 89.50 & 88.21 & 87.67 & 80.95 & 93.44 & 87.19 & 75.40 \\
& \textbf{Ours} & \textbf{96.00} & \textbf{97.18} & \textbf{95.32} & \textbf{87.88} & \textbf{100} & \textbf{100} & \textbf{91.06} \\
\bottomrule
\end{tabular}%
}
\end{center}

\begin{center}
\captionof{table}{Effect of Data Preprocessing on State-of-the-Art Algorithm Performance: Original vs Improved UP-Fall Dataset Comparison. Bold values indicate highest performance.}
\label{tab10}
\small
\begin{tabular}{l|l|c|c|c|c}
\hline
\textbf{Model} & \textbf{Dataset} & \textbf{Accuracy} & \textbf{Precision} & \textbf{Recall} & \textbf{Specificity} \\
 &  & (\%) & (\%) & (\%) & (\%) \\ \hline
CNN\citep{ha2024fall} & Original Data & 92.00 & 87.71 & 94.24& 84.21\\ 
 & Improved Data & \textbf{96.00} & \textbf{97.18} & \textbf{100} & \textbf{87.88} \\ \hline
LSTM\citep{castro2023fall} & Original Data  & 80.00 & 71.66 & 77.71 & \textbf{90.91} \\ 
 & Improved Data & \textbf{85.00} & \textbf{81.00} & \textbf{85.27} & 88.33\\ \hline
BiLSTM\citep{csengul2022deep} & Original Data & 88.10 & 83.77 & 87.83 &  80.30 \\ 
 & Improved Data & \textbf{93.50} & \textbf{95.58} & \textbf{100} &\textbf{89.25}\\ \hline
STGCN\citep{abdo2023human} & Original Data & 72.00 & 69.95 &  73.85 &  68.57\\ 
 & Improved Data & \textbf{87.43} & \textbf{84.74} &  \textbf{90.86} &  \textbf{91.18}\\ \hline
\textbf{Ours} & Original Data  & 95.00 & 96.43 & 100 & 85.71\\ 
 & Improved Data & \textbf{97.00} & \textbf{97.86} & \textbf{100} & \textbf{91.71}\\ \hline
\end{tabular}
\end{center}

\subsection{Cross-Dataset Validation on UMAFall Dataset}

To assess the generalization capability of our approach across different datasets, we evaluated all algorithms' performance on the UMAFall dataset, which was completely distinct from the UP-Fall dataset used for training. This cross-dataset evaluation demonstrates how our preprocessing methodology affects algorithm performance when applied consistently across both training and testing datasets from different sources.
All algorithms were trained on our improved UP-Fall dataset (with our preprocessing technique applied) and tested on the UMAFall dataset that was also processed using our preprocessing technique. This approach enables evaluation of algorithm performance when both training and testing data use our consistent data processing methodology across different datasets.
The results in Table \ref{tab11} demonstrate the effectiveness of our preprocessing approach for cross-dataset generalization. Our proposed method achieved the highest overall performance with 94\% accuracy, demonstrating it generalization capability across different datasets. The consistent performance across all state-of-the-art algorithms when using our preprocessing approach highlights the value and transferability of our data enhancement techniques.
These results validate both the generalizability of our impact detection method across different datasets and the effectiveness of our preprocessing technique in enhancing cross-dataset performance. This suggests that our preprocessing methodology provides significant benefits when applied consistently across different datasets, making it a valuable contribution for the impact fall detection research community.

\begin{center}
\captionof{table}{Cross-Dataset Performance Evaluation: Training on Improved UP-Fall, Testing on Preprocessed UMAFall Dataset. Bold values indicate highest performance.}
\label{tab11}
\begin{tabular}{l|c|c|c|c}
\hline
\textbf{Model} & \textbf{Accuracy} & \textbf{Precision} & \textbf{Recall} & \textbf{Specificity} \\
 & (\%) & (\%) & (\%) & (\%) \\ 
\hline
CNN~\cite{ha2024fall} & 91.33 & 90.88 & 95.59 & 82.29 \\ 
\hline
LSTM~\cite{lau2022fall} & 83.67 & 91.43 & 97.34 & 84.38 \\ 
\hline
BiLSTM~\cite{csengul2022deep} & 93.33 & 93.29 & 97.06 & 85.45 \\ 
\hline
STGCN~\cite{abdo2023human} & 92.00 & 91.43 & 95.59 & 84.38 \\ 
\hline
\textbf{Ours} & \textbf{94.00} & \textbf{94.00} & \textbf{97.34} & \textbf{87.04} \\ 
\hline
\end{tabular}
\end{center}

\subsection{\centering Visual Comparison between our Method and the state of the art Fall Detection Algorithms}

In this section, we conduct further comparisons between our proposed method and state-of-the-art fall detection algorithms.
The STGCN baseline architecture, demonstrated promising results, as previously presented in Table \ref{tab8}. However, despite the overall satisfactory performance, upon visual examination, it became apparent that the impact identified by the current STGCN algorithm \citep{abdo2023human} did not consistently align with impactful events visually inspected as shown in Figure \ref{stgcn_test}.(a).
This emphasizes the need for further refinement in algorithms to not only identify falls but also precisely recognize impactful events within those falls. In contrast, our proposed STGCN-GRU-BiLSTM algorithm achieved a good performance. Moreover,  upon visual inspection, we could correctly identify impact as shown in Figure \ref{stgcn_test}.(b). This indicates the effectiveness of our method in accurately detecting impactful events within falls, contributing to the development of a more reliable fall detection system.

\begin{center}
\includegraphics[width=14cm]{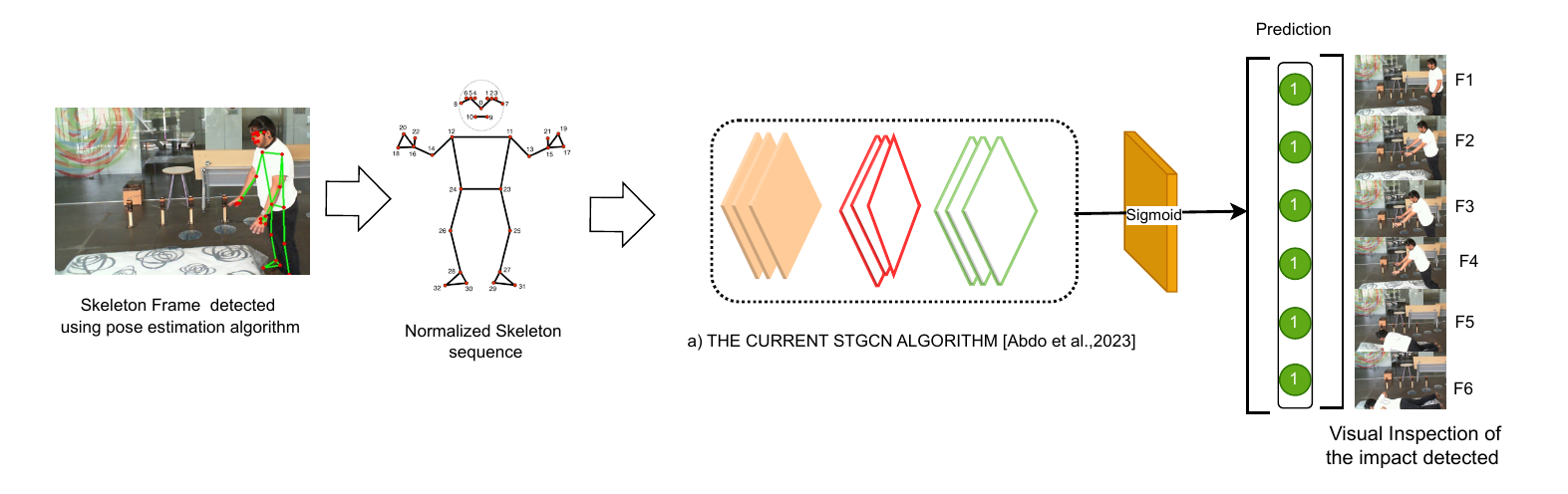}
\includegraphics[width=14cm]{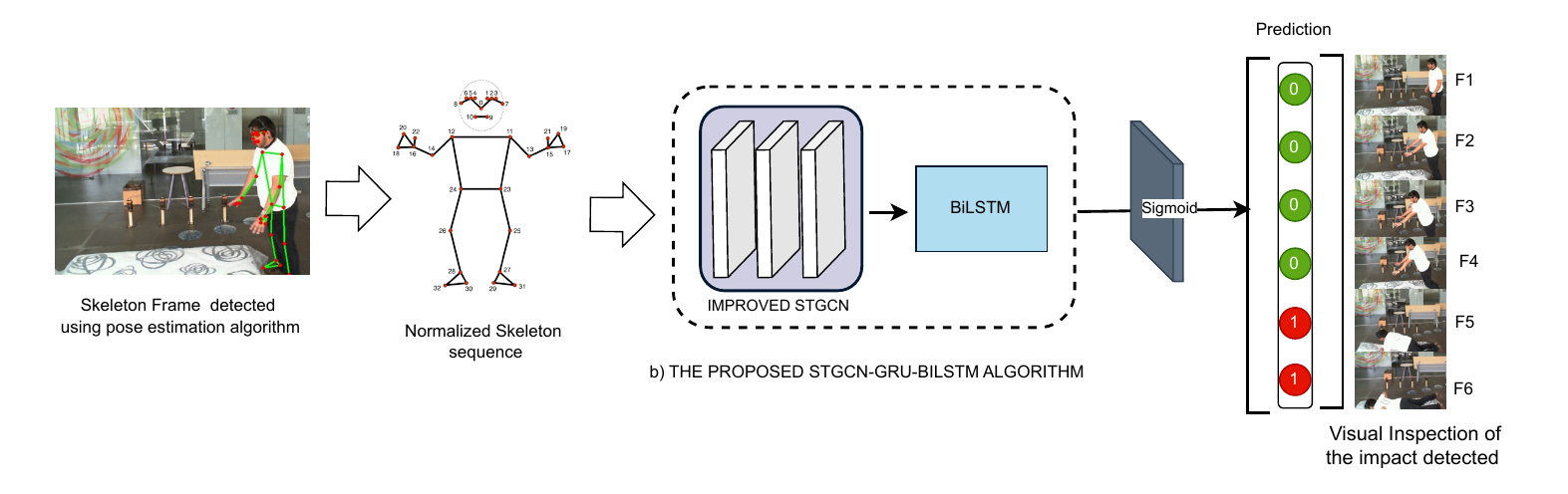}
\captionof{figure}{Visual Comparison of the Impact Detected using the current STGCN \citep{abdo2023human} trained with the initial UP-Fall dataset with our Proposed STGCN-GRU-BiLSTM trained with the improved UP-Fall dataset; upon visual inspection a) the current STGCN failed to accurately detect impact where our proposed STGCN-GRU-BiLSTM b) accurately detected impact (F5). F1..F6 refer to the sequence frame}
\label{stgcn_test}
\end{center}

\subsection{Role of Body Joints in Impact Detection}

To investigate the role of body joints in detecting impact within fall events, we conducted a comprehensive analysis focusing on joint activation patterns associated with impact across diverse subjects performing similar fall scenarios, as illustrated in Figure~\ref{body_joint}. Our primary objective was to determine whether the body joints relevant to impact detection remain consistent across different individuals, regardless of their fall dynamics. We employed an attention vector mechanism within our STGCN-GRU-BiLSTM framework to dynamically prioritize joints based on their relevance to impact during falls. The attention vector \(Va\) operates by leveraging the adjacency matrix, which represents connections between joints, to guide the graph convolution process. This method enables dynamic weighting of joint connections, emphasizing those joints that typically experience higher impact forces during falls. Our analysis enables the model to focus on joints that are likely initial points of contact with the ground or experience significant forces during falls. As shown in Figure~\ref{body_joint}, our findings identify several critical joints with consistently higher attention scores: the head, right shoulder, left elbow, right wrist, left hip, and right knee. These joints represent the most probable initial contact points or areas experiencing significant forces during impact events. The consistent activation of these specific joints across various subjects and fall scenarios underscores their crucial role in impact detection. This approach facilitates identification of the most impact-relevant joints while enhancing understanding of fall dynamics across varied scenarios. The insights gained from attention vector analysis provide valuable foundations for developing targeted injury prevention strategies, particularly for protecting vulnerable body parts such as the hips, knees, and head during falls.These findings establish the critical importance of specific body joints in impact detection, providing essential groundwork for refining fall detection algorithms and informing evidence-based injury prevention strategies.

\begin{center}
\includegraphics[width=13.8cm]{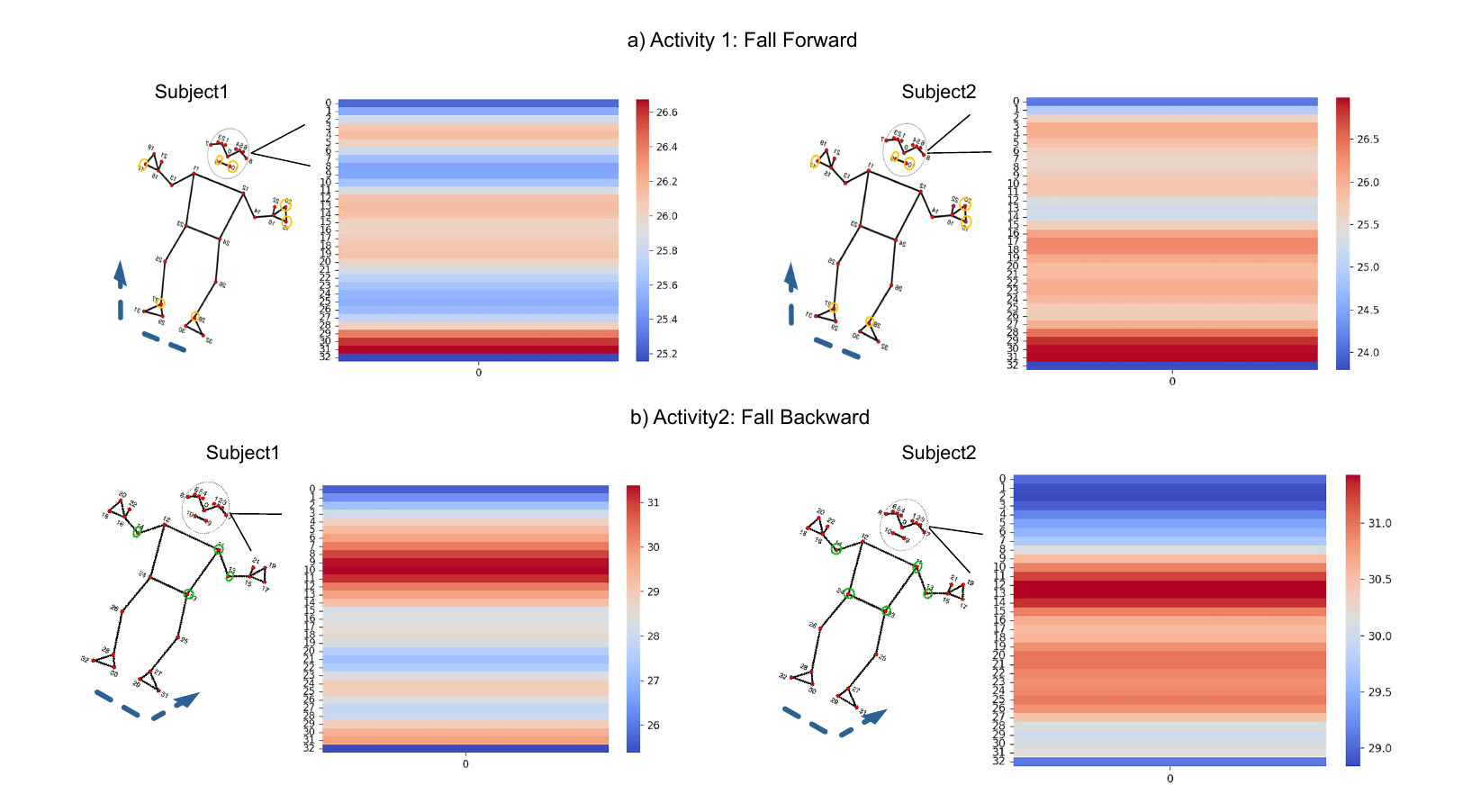}
\captionof{figure}{Visualization of the average attention vector and the role of body joints for two different falls on two different subjects; The two lines on the head of subjects refer to the fall direction}
\label{body_joint}
\end{center}
% [scale=0.6]
% [width=15cm]
% [scale=0.25]

\subsection{Analysis of Impact Dynamics Through BiLSTM}

To investigate the potential of BiLSTM layers within our model for characterizing human movement dynamics, particularly in detecting impact within fall events, we conducted a comprehensive analysis of the learned feature representations. These BiLSTM layers serve as critical components for capturing complex spatio-temporal features inherent in human motion data. By examining their outputs, we aimed to analyze and quantify impact characteristics while gaining deeper insights into fall dynamics.
The analysis of BiLSTM outputs to identify patterns indicative of impact events provides a strategic approach to understanding fall dynamics. After training our model, we conducted tests using fixed-length inputs (60, 90, and 180 frames) of the same test data, representing falls at different velocities. We then extracted the latent feature representations from the BiLSTM layer for each fixed-length sequence. These features provided valuable metrics characterizing impact strength, duration, and frequency, enabling effective quantification of impact events.
To visualize these high-dimensional features, we employed autoencoders~\cite{balin2019concrete} to project the multidimensional impact data into a lower-dimensional space, revealing how fall velocity influences feature representations as illustrated in 
 Figure~\ref{tsne}. This visualization demonstrates how variations in fall speed affect the clustering of feature representations and confirms whether the model differentiates between falls of varying velocities.
The resulting plot distinctly separates "Fall Forward" and "Fall Backward" scenarios with clear clustering based on fall speed. Falls with longer duration (more frames) form distinct sub-clusters that are as distinguishable as those with shorter duration. These results confirm that the model generalizes effectively to different fall scenarios, successfully characterizing impact events based on both speed and type. This capability is crucial for evaluating model robustness in real-world scenarios with varying fall velocities.

\begin{center}
\includegraphics[width=14cm]{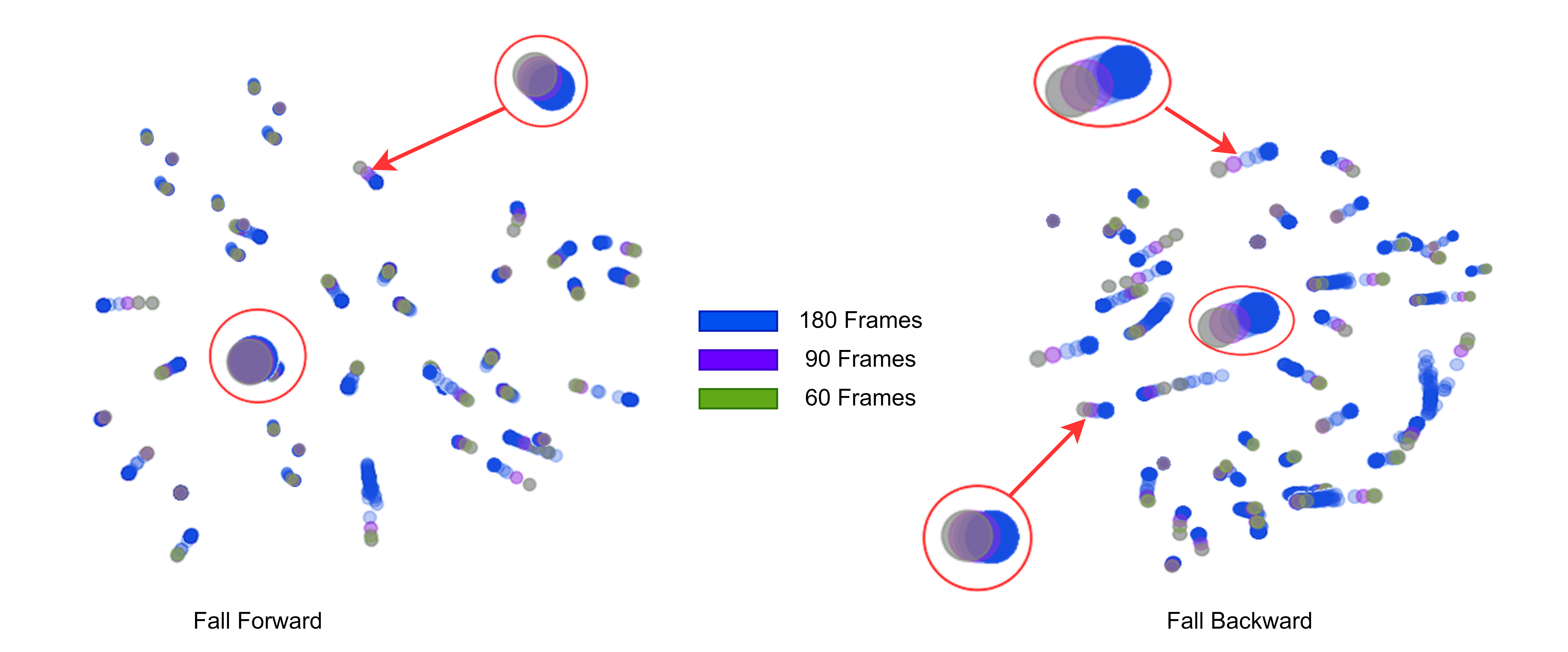}
\captionof{figure}{Autoencoder \citep{balin2019concrete} visualization of impact dynamics for two types of falls (Forward and Backward) from the UP-Fall Dataset. Our model identifies consistent features from various fixed lengths sequences of the same fall.}
\label{tsne}
\end{center}

\subsection{Impact Detection Under Partial Occlusion}

Real-world deployment of impact detection systems often encounters scenarios where subjects are partially occluded by environmental objects during fall events. To evaluate our model's robustness in such conditions, we analyzed impact detection performance under partial occlusion scenarios. The full visibility baseline represents our experimental results reported throughout this paper.

\begin{table}[h]
\centering
\caption{Impact Detection Performance Under Partial Occlusion Scenarios}
\label{tab:occlusion_performance}
\begin{tabular}{lcc}
\hline
\textbf{Visibility Condition} & \textbf{Active Joints} & \textbf{Accuracy (\%)} \\
\hline
Full Visibility (Baseline) & 33/33 (100\%) & 97.50 \\
Upper Body Visible & 23/33 (70\%) & 95.20 \\
Lower Body Visible & 10/33 (30\%) & 76.60 \\
\hline
\end{tabular}
\end{table}

Our STGCN-GRU-BiLSTM architecture demonstrates robust impact detection performance under partial occlusion through graph convolution operations that leverage anatomical joint relationships, bidirectional LSTM processing that maintains temporal context across occlusion events, and attention mechanisms that dynamically focus on visible joints.
As shown in Table~\ref{tab:occlusion_performance}, when the upper body remains visible (common scenarios include subjects falling behind desks, tables, or low barriers where 23 of 33 joints are active), the model maintains 95.2\% accuracy in impact detection. When only the lower body is visible (high barriers or obstacles obscuring the upper torso, with only 10 of 33 joints active), performance decreases to 76.6\% due to reduced joint information and absence of critical upper body landmarks for impact identification.
These results demonstrate that our method effectively identifies impact moments within fall events when upper body visibility is maintained, while showing limitations under severe occlusion conditions.

\begin{center}
\captionof{table}{Performance and Resource Analysis of Models with Improved 3D Skeleton Data (300 Epochs)}
\label{tab12}
\footnotesize
\begin{tabular}{lcccc}
\hline
\textbf{Models} & \textbf{Training Time} & \textbf{Testing Time} & \textbf{Trainable Parameters} & \textbf{VRAM} \\
\hline
STGCN & 56 min & 17 sec & 344,809 & 86.38 MB \\
STGCN ConvLSTM & 3h 40 min & 27 sec & 858,705 & 193.53 MB \\
STGCN-GRU & 39 min & 8 sec & 585,999 & 124.92 MB \\
\textbf{Ours} & 41 min & 12 sec & 1,417,524 & 212.65 MB \\
\hline
\end{tabular}
\end{center}

\subsection{Computational and Memory Cost Analysis}

The computational efficiency of our proposed method was evaluated on the improved 3D skeleton UP-Fall dataset using a GeForce GTX 3080 Ti with 16GB of RAM. While computational efficiency was not our primary objective, as our focus was developing an accurate impact detection algorithm for fall events, we present these metrics in Table~\ref{tab12} to demonstrate the trade-offs between computational demands and memory consumption across different model architectures under identical training conditions.
The baseline STGCN with average pooling achieved a training time of 56 minutes and testing time of 17 seconds, utilizing 344,809 parameters and 86.38 MB of VRAM. This configuration provides computational efficiency but with limited temporal modeling capabilities. The STGCN-ConvLSTM variant, incorporating more sophisticated temporal processing, required substantially higher computational resources with a training time of 3 hours 40 minutes, testing time of 27 seconds, 858,705 parameters, and 193.53 MB of VRAM usage.
Significant efficiency improvements were achieved with the STGCN-GRU architecture, reducing training time to 39 minutes and testing time to 8 seconds while utilizing 585,999 parameters and 124.92 MB of VRAM. This demonstrates the computational advantages of GRU units for temporal sequence modeling compared to ConvLSTM approaches.
Our final STGCN-GRU-BiLSTM model achieved a training time of 41 minutes and testing time of 12 seconds (6 seconds per sequence), utilizing 1,417,524 parameters and 212.65 MB of VRAM. Despite having the highest parameter count and memory usage among compared models, this architecture benefits from BiLSTM's bidirectional temporal dependency modeling, which significantly enhances impact detection accuracy and model robustness.
The strategic integration of GRUs and BiLSTMs in our architecture optimizes the trade-off between computational efficiency and detection performance. While GRUs provide efficient temporal processing, BiLSTMs enhance bidirectional temporal analysis, improving overall impact detection accuracy. Notably, despite having fewer parameters than our final model, the STGCN-ConvLSTM required significantly longer training time, highlighting the efficiency advantages of our architectural design.
These results demonstrate that our approach achieves superior impact detection performance while maintaining computational feasibility for practical applications requiring timely and accurate fall event detection.

\section{Conclusion}
\label{sec:Conclusion}
This study presents a novel approach to fall detection that focuses on detecting the precise moment of impact within fall events. We propose a STGCN-GRU-BiLSTM architecture that integrates Gated Recurrent Unit (GRU) and Bidirectional Long Short-Term Memory (BiLSTM) layers onto a Spatio-Temporal Graph Convolutional Network (STGCN) baseline architecture. The GRU layers enhance processing efficiency while maintaining accurate temporal analysis of joint movements, while BiLSTM layers enable bidirectional temporal modeling to capture impact events accurately across variable-length sequences. Our ablation study validates the unique contribution of each component, demonstrating efficient spatio-temporal analysis capabilities.
By leveraging 3D joint skeleton data, our model successfully distinguishes between false alarms and actual impact events, significantly reducing false positive rates. Evaluated on the improved 3D skeleton UP-Fall dataset, our model demonstrated good performance, achieving 97.50\% accuracy for impact detection. The model particularly excels in detecting impacts during backward falls while standing (96.50\% accuracy) and falls while sitting (97.50\% accuracy), highlighting its practical applicability across diverse fall scenarios. 
The cross-dataset evaluation on the UMAFall dataset, preprocessed using our labeling methodology, demonstrated the model's generalization capability for previously unseen fall scenarios while maintaining consistent performance. Comparative analysis of state-of-the-art algorithms on both original and preprocessed datasets reveals that our preprocessing approach significantly enhances performance metrics across all tested algorithms. These results validate both our labeling methodology's effectiveness and its potential to improve fall detection system quality. This contributes to better healthcare resource allocation and more timely interventions for elderly populations at risk of fall-related injuries.

\section{Future Scope}
\label{sec:FutureScope}
This study opens promising avenues for future research and practical deployment. The demonstrated robustness under partial occlusion conditions, with maintained performance of 95.2\% accuracy when upper-body remains visible, provides a strong foundation for real-world applications. Future work can leverage these promising occlusion results by integrating our approach into a comprehensive pipeline that incorporates person segmentation modules. Hence, this architecture would first detect and segment individual persons in the scene, then apply our impact detection method to each resulting region of interest (ROI). Such integration would enable simultaneous multi-person monitoring while maintaining the high temporal precision demonstrated in our single-person analysis.
Second, extending validation to real-world elderly care environments represents a natural progression, building upon our current framework's demonstrated accuracy of 97.5\% on simulated data. Future research will focus on collecting and analyzing real-world fall data from actual elderly populations to validate our approach's effectiveness. This real-world validation will provide critical insights into system performance across diverse elderly demographics and care environments where partial occlusion is common.
Finally, exploring knowledge distillation and model compression techniques could enable the development of computationally efficient variants suitable for deployment on resource-constrained healthcare devices while maintaining high accuracy. These developments together can significantly improve both the scalability and practicality of impact detection systems for comprehensive healthcare monitoring.

% \noindent\textbf{Funding} This research was supported by CESI and its research laboratory CESI LINEACT, France
\noindent\textbf{Funding} This research was supported by CESI Engineering School through the CESI LINEACT laboratory in France.

\bibliography{sn-bibliography}

@article{martinez2019up,
  title={UP-fall detection dataset: A multimodal approach},
  author={Mart{\'\i}nez-Villase{\~n}or, Lourdes and Ponce, Hiram and Brieva, Jorge and Moya-Albor, Ernesto and N{\'u}{\~n}ez-Mart{\'\i}nez, Jos{\'e} and Pe{\~n}afort-Asturiano, Carlos},
  journal={Sensors},
  volume={19},
  number={9},
  pages={1988},
  year={2019},
  publisher={MDPI}
}

@inproceedings{chi2023prefallkd,
  title={Prefallkd: Pre-Impact Fall Detection Via CNN-ViT Knowledge Distillation},
  author={Chi, Tin-Han and Liu, Kai-Chun and Hsieh, Chia-Yeh and Tsao, Yu and Chan, Chia-Tai},
  booktitle={ICASSP 2023-2023 IEEE International Conference on Acoustics, Speech and Signal Processing (ICASSP)},
  pages={1--5},
  year={2023},
  organization={IEEE}
}

@article{bourke2007evaluation,
  title={Evaluation of a threshold-based tri-axial accelerometer fall detection algorithm},
  author={Bourke, Alan K and O’brien, JV and Lyons, Gerard M},
  journal={Gait \& posture},
  volume={26},
  number={2},
  pages={194--199},
  year={2007},
  publisher={Elsevier}
}

@inproceedings{koffi2023machine,
  title={Machine Learning and Feature Ranking for Impact Fall Detection Event Using Multisensor Data},
  author={Koffi, Tresor Y and Mourchid, Youssef and Hindawi, Mohammed and Dupuis, Yohan},
  booktitle={2023 IEEE 25th International Workshop on Multimedia Signal Processing (MMSP)},
  pages={1--6},
  year={2023},
  organization={IEEE}
}

@article{lin2020fall,
  title={Fall monitoring for the elderly using wearable inertial measurement sensors on eyeglasses},
  author={Lin, Chih-Lung and Chiu, Wen-Ching and Chen, Fu-Hsing and Ho, Yuan-Hao and Chu, Ting-Ching and Hsieh, Ping-Hsiao},
  journal={IEEE Sensors Letters},
  volume={4},
  number={6},
  pages={1--4},
  year={2020},
  publisher={IEEE}
}

@inproceedings{wasi2022machine,
  title={Machine Learning Algorithm for Fall Classification Using Wearable Device},
  author={Wasi, Md Wasif Islam and Dziyauddin, Rudzidatul Akmam and Amir, Nur Izdihar Muhd and Ahmad, Robiah},
  booktitle={2022 IEEE Symposium on Future Telecommunication Technologies (SOFTT)},
  pages={62--66},
  year={2022},
  organization={IEEE}
}

@article{kausar2022automated,
  title={Automated machine learning based elderly fall detection classification},
  author={Kausar, Firdous and Awadalla, Medhat and Mesbah, Mostefa and AlBadi, Taif},
  journal={Procedia Computer Science},
  volume={203},
  pages={16--23},
  year={2022},
  publisher={Elsevier}
}

@inproceedings{shen2019fall,
  title={Fall detection system based on deep learning and image processing in cloud environment},
  author={Shen, Leixian and Zhang, Qingyun and Cao, Guoxu and Xu, He},
  booktitle={Complex, Intelligent, and Software Intensive Systems: Proceedings of the 12th International Conference on Complex, Intelligent, and Software Intensive Systems (CISIS-2018)},
  pages={590--598},
  year={2019},
  organization={Springer}
}

@article{csengul2022deep,
  title={Deep learning based fall detection using smartwatches for healthcare applications},
  author={{\c{S}}eng{\"u}l, G{\"o}khan and Karakaya, Murat and Misra, Sanjay and Abayomi-Alli, Olusola O and Dama{\v{s}}evi{\v{c}}ius, Robertas},
  journal={Biomedical Signal Processing and Control},
  volume={71},
  pages={103242},
  year={2022},
  publisher={Elsevier}
}

@article{grishchenko2022blazepose,
  title={Blazepose ghum holistic: Real-time 3d human landmarks and pose estimation},
  author={Grishchenko, Ivan and Bazarevsky, Valentin and Zanfir, Andrei and Bazavan, Eduard Gabriel and Zanfir, Mihai and Yee, Richard and Raveendran, Karthik and Zhdanovich, Matsvei and Grundmann, Matthias and Sminchisescu, Cristian},
  journal={arXiv preprint arXiv:2206.11678},
  year={2022}
}

@article{yu2020novel,
  title={A novel hybrid deep neural network to predict pre-impact fall for older people based on wearable inertial sensors},
  author={Yu, Xiaoqun and Qiu, Hai and Xiong, Shuping},
  journal={Frontiers in bioengineering and biotechnology},
  volume={8},
  pages={63},
  year={2020},
  publisher={Frontiers Media SA}
}

@article{keskes2021vision,
  title={Vision-based fall detection using st-gcn},
  author={Keskes, Oussema and Noumeir, Rita},
  journal={IEEE Access},
  volume={9},
  pages={28224--28236},
  year={2021},
  publisher={IEEE}
}

@article{balasubramaniam2023modified,
  title={A modified LeNet CNN for breast cancer diagnosis in ultrasound images},
  author={Balasubramaniam, Sathiyabhama and Velmurugan, Yuvarajan and Jaganathan, Dhayanithi and Dhanasekaran, Seshathiri},
  journal={Diagnostics},
  volume={13},
  number={17},
  pages={2746},
  year={2023},
  publisher={MDPI}
}

@article{putra2017event,
  title={An event-triggered machine learning approach for accelerometer-based fall detection},
  author={Putra, I Putu Edy Suardiyana and Brusey, James and Gaura, Elena and Vesilo, Rein},
  journal={Sensors},
  volume={18},
  number={1},
  pages={20},
  year={2017},
  publisher={MDPI}
}

@article{droghini2017combined,
  title={A combined one-class SVM and template-matching approach for user-aided human fall detection by means of floor acoustic features},
  author={Droghini, Diego and Ferretti, Daniele and Principi, Emanuele and Squartini, Stefano and Piazza, Francesco and others},
  journal={Computational intelligence and neuroscience},
  volume={2017},
  year={2017},
  publisher={Hindawi}
}

@article{jeong2022fall,
  title={Fall Detection System Based on Simple Threshold Method and Long Short-Term Memory: Comparison with Hidden Markov Model and Extraction of Optimal Parameters},
  author={Jeong, Seung Su and Kim, Nam Ho and Yu, Yun Seop},
  journal={Applied Sciences},
  volume={12},
  number={21},
  pages={11031},
  year={2022},
  publisher={MDPI}
}

@inproceedings{xu2019cnn,
  title={CNN-LSTM combined network for IoT enabled fall detection applications},
  author={Xu, Jun and He, Zunwen and Zhang, Yan},
  booktitle={Journal of Physics: Conference Series},
  volume={1267},
  number={1},
  pages={012044},
  year={2019},
  organization={IOP Publishing}
}

@article{vujovic2021classification,
  title={Classification model evaluation metrics},
  author={Vujovi{\'c}, {\v{Z}} and others},
  journal={International Journal of Advanced Computer Science and Applications},
  volume={12},
  number={6},
  pages={599--606},
  year={2021}
}

@article{yang2022fall,
  title={Fall detection system based on infrared array sensor and multi-dimensional feature fusion},
  author={Yang, Yi and Yang, Honglei and Liu, Zhixin and Yuan, Yazhou and Guan, Xinping},
  journal={Measurement},
  volume={192},
  pages={110870},
  year={2022},
  publisher={Elsevier}
}

@inproceedings{noury2007fall,
  title={Fall detection-principles and methods},
  author={Noury, Norbert and Fleury, Anthony and Rumeau, Pierre and Bourke, Alan K and Laighin, GO and Rialle, Vincent and Lundy, Jean-Eric},
  booktitle={2007 29th annual international conference of the IEEE engineering in medicine and biology society},
  pages={1663--1666},
  year={2007},
  organization={IEEE}
}

@article{mubashir2013survey,
  title={A survey on fall detection: Principles and approaches},
  author={Mubashir, Muhammad and Shao, Ling and Seed, Luke},
  journal={Neurocomputing},
  volume={100},
  pages={144--152},
  year={2013},
  publisher={Elsevier}
}

@article{nooruddin2020iot,
  title={An IoT based device-type invariant fall detection system},
  author={Nooruddin, Sheikh and Islam, Md Milon and Sharna, Falguni Ahmed},
  journal={Internet of Things},
  volume={9},
  pages={100130},
  year={2020},
  publisher={Elsevier}
}

@article{ren2019research,
  title={Research of fall detection and fall prevention technologies: A systematic review},
  author={Ren, Lingmei and Peng, Yanjun},
  journal={IEEE Access},
  volume={7},
  pages={77702--77722},
  year={2019},
  publisher={IEEE}
}

@article{kottari2019real,
  title={Real-time fall detection using uncalibrated fisheye cameras},
  author={Kottari, Konstantina N and Delibasis, Konstantinos K and Maglogiannis, Ilias G},
  journal={IEEE transactions on cognitive and developmental systems},
  volume={12},
  number={3},
  pages={588--600},
  year={2019},
  publisher={IEEE}
}

@article{espinosa2020application,
  title={Application of convolutional neural networks for fall detection using multiple cameras},
  author={Espinosa, Ricardo and Ponce, Hiram and Guti{\'e}rrez, Sebasti{\'a}n and Mart{\'\i}nez-Villase{\~n}or, Lourdes and Brieva, Jorge and Moya-Albor, Ernesto},
  journal={Challenges and Trends in Multimodal Fall Detection for Healthcare},
  pages={97--120},
  year={2020},
  publisher={Springer}
}

@article{chen2020fall,
  title={Fall detection based on key points of human-skeleton using openpose},
  author={Chen, Weiming and Jiang, Zijie and Guo, Hailin and Ni, Xiaoyang},
  journal={Symmetry},
  volume={12},
  number={5},
  pages={744},
  year={2020},
  publisher={MDPI}
}

@article{islam2020deep,
  title={Deep learning based systems developed for fall detection: a review},
  author={Islam, Md Milon and Tayan, Omar and Islam, Md Repon and Islam, Md Saiful and Nooruddin, Sheikh and Kabir, Muhammad Nomani and Islam, Md Rabiul},
  journal={IEEE Access},
  volume={8},
  pages={166117--166137},
  year={2020},
  publisher={IEEE}
}

@article{vyas2019pose,
  title={Pose estimation and action recognition in sports and fitness},
  author={Vyas, Parth},
  year={2019}
}

@article{kim2020implementation,
  title={Implementation of a real-time fall detection system for elderly Korean farmers using an insole-integrated sensing device},
  author={Kim, Insoo and Lee, Kyung-Suk and Kim, Kyungran and Kim, Kyungsu and Chae, Hye-Seon and Kim, Hyo-Cher},
  journal={Instrumentation Science \& Technology},
  volume={48},
  number={1},
  pages={22--42},
  year={2020},
  publisher={Taylor \& Francis}
}

@article{lugaresi2019mediapipe,
  title={Mediapipe: A framework for building perception pipelines},
  author={Lugaresi, Camillo and Tang, Jiuqiang and Nash, Hadon and McClanahan, Chris and Uboweja, Esha and Hays, Michael and Zhang, Fan and Chang, Chuo-Ling and Yong, Ming Guang and Lee, Juhyun and others},
  journal={arXiv preprint arXiv:1906.08172},
  year={2019}
}

@article{kim2019machine,
  title={Machine learning-based pre-impact fall detection model to discriminate various types of fall},
  author={Kim, Tae Hyong and Choi, Ahnryul and Heo, Hyun Mu and Kim, Kyungran and Lee, Kyungsuk and Mun, Joung Hwan},
  journal={Journal of biomechanical engineering},
  volume={141},
  number={8},
  pages={081010},
  year={2019},
  publisher={American Society of Mechanical Engineers}
}

@article{bourke2008threshold,
  title={A threshold-based fall-detection algorithm using a bi-axial gyroscope sensor},
  author={Bourke, Alan K and Lyons, Gerald M},
  journal={Medical engineering \& physics},
  volume={30},
  number={1},
  pages={84--90},
  year={2008},
  publisher={Elsevier}
}

@article{bazarevsky2020blazepose,
  title={Blazepose: On-device real-time body pose tracking},
  author={Bazarevsky, Valentin and Grishchenko, Ivan and Raveendran, Karthik and Zhu, Tyler and Zhang, Fan and Grundmann, Matthias},
  journal={arXiv preprint arXiv:2006.10204},
  year={2020}
}

@phdthesis{martinez2019openpose,
  title={Openpose: Whole-body pose estimation},
  author={Mart{\i}nez, Gin{\'e}s Hidalgo},
  year={2019},
  school={Carnegie Mellon University Pittsburgh, PA, USA}
}

@article{casilari2017umafall,
  title={Umafall: A multisensor dataset for the research on automatic fall detection},
  author={Casilari, Eduardo and Santoyo-Ram{\'o}n, Jose A and Cano-Garc{\'\i}a, Jose M},
  journal={Procedia Computer Science},
  volume={110},
  pages={32--39},
  year={2017},
  publisher={Elsevier}
}

@article{wang2023review,
  title={Review of GrabCut in image processing},
  author={Wang, Zhaobin and Lv, Yongke and Wu, Runliang and Zhang, Yaonan},
  journal={Mathematics},
  volume={11},
  number={8},
  pages={1965},
  year={2023},
  publisher={MDPI}
}

@article{wang2023fall,
  title={Fall detection with a non-intrusive and first-person vision approach},
  author={Wang, Xueyi and Talavera, Estefan{\'\i}a and Karastoyanova, Dimka and Azzopardi, George},
  journal={IEEE Sensors Journal},
  year={2023},
  publisher={IEEE}
}

@article{chicco2020advantages,
  title={The advantages of the Matthews correlation coefficient (MCC) over F1 score and accuracy in binary classification evaluation},
  author={Chicco, Davide and Jurman, Giuseppe},
  journal={BMC genomics},
  volume={21},
  pages={1--13},
  year={2020},
  publisher={Springer}
}

@article{ximenes2024impact,
  title={Impact of educational intervention in the perception of hospitalised patients about the risk of falling and associated factors},
  author={Ximenes, Maria Aline Moreira and Oliveira, Ingrid Kelly Morais and Cavalcante, Francisco Marcelo Leandro and Neto, Nelson Miguel Galindo and Caetano, Joselany {\'A}fio and Barros, L{\'\i}via Moreira},
  journal={Authorea Preprints},
  year={2024},
  publisher={Authorea}
}

@article{wangfusion,
  title={Fusion of Machine Learning and Threshold-Based Approaches for Fall Detection in Healthcare Using Inertial Sensors},
  author={Wang, Ya and Sarvari, Peiman Alipour and Khadraoui, Djamel}
}

@article{mourchid2016image,
  title={Image segmentation based on community detection approach},
  author={Mourchid, Youssef and El Hassouni, Mohammed and Cherifi, Hocine},
  journal={International Journal of Computer Information Systems and Industrial Management Applications},
  volume={8},
  pages={195--204},
  year={2016}
}

@inproceedings{lafhel2021movie,
  title={Movie script similarity using multilayer network portrait divergence},
  author={Lafhel, Majda and Cherifi, Hocine and Renoust, Benjamin and El Hassouni, Mohammed and Mourchid, Youssef},
  booktitle={Complex Networks \& Their Applications IX: Volume 1, Proceedings of the Ninth International Conference on Complex Networks and Their Applications COMPLEX NETWORKS 2020},
  pages={284--295},
  year={2021},
  organization={Springer}
}

@inproceedings{cherifi2017complex,
  title={Complex Networks and Their Applications VI},
  author={Cherifi, Chantal and Cherifi, Hocine and Karsai, M{\'a}rton and Musolesi, Mirco},
  booktitle={Proceedings of Complex Networks 2017 The Sixth International Conference on Complex Networks and Their Applications. Berlin: Springer},
  year={2017},
  organization={Springer}
}

@article{mourchid2023d,
  title={D-STGCNT: A Dense Spatio-Temporal Graph Conv-GRU Network based on transformer for assessment of patient physical rehabilitation},
  author={Mourchid, Youssef and Slama, Rim},
  journal={Computers in Biology and Medicine},
  volume={165},
  pages={107420},
  year={2023},
  publisher={Elsevier}
}

@article{eltahir2023deep,
  title={Deep Transfer Learning-Enabled Activity Identification and Fall Detection for Disabled People.},
  author={Eltahir, Majdy M and Yousif, Adil and Alrowais, Fadwa and Nour, Mohamed K and Marzouk, Radwa and Dafaalla, Hatim and Hassan Elnour, Asma Abbas and Aziz, Amira Sayed A and Hamza, Manar Ahmed},
  journal={Computers, Materials \& Continua},
  volume={75},
  number={2},
  year={2023}
}

@article{lau2022fall,
  title={Fall detection and motion analysis using visual approaches},
  author={Lau, Xin Lin and Connie, Tee and Goh, Michael Kah Ong and Lau, Siong Hoe},
  journal={International Journal of Technology},
  volume={13},
  number={6},
  pages={1173--1182},
  year={2022},
  publisher={IJTech}
}

@article{abdo2023human,
  title={Human Fall Detection Using Spatial Temporal Graph Convolutional Networks.},
  author={Abdo, Hadeer Atef and Amin, Khalid and Hamad, Ahmed Mahmoud},
  journal={IJCI. International Journal of Computers and Information},
  volume={10},
  number={2},
  pages={80--98},
  year={2023},
  publisher={Minufiya University; Faculty of Computers and Information}
}

@article{ha2024fall,
  title={Fall detection using mixtures of convolutional neural networks},
  author={Ha, Thao V and Nguyen, Hoang M and Thanh, Son H and Nguyen, Binh T},
  journal={Multimedia Tools and Applications},
  volume={83},
  number={6},
  pages={18091--18118},
  year={2024},
  publisher={Springer}
}

@inproceedings{castro2023fall,
  title={Fall Detection with LSTM and Attention Mechanism.},
  author={Castro, Francesco and Dentamaro, Vincenzo and Gattulli, Vincenzo and Impedovo, Donato},
  booktitle={WAMWB@ MobileHCI},
  pages={37--50},
  year={2023}
}

@article{ma2024application,
  title={Application of Dual-Stage Attention Temporal Convolutional Networks in Gas Well Production Prediction},
  author={Ma, Xianlin and Zhang, Long and Zhan, Jie and Chang, Shilong},
  journal={Mathematics},
  volume={12},
  number={24},
  pages={3896},
  year={2024},
  publisher={MDPI}
}

@article{kibet2024sudden,
  title={Sudden Fall Detection of Human Body Using Transformer Model},
  author={Kibet, Duncan and So, Min Seop and Kang, Hahyeon and Han, Yongsu and Shin, Jong-Ho},
  journal={Sensors},
  volume={24},
  number={24},
  pages={8051},
  year={2024},
  publisher={MDPI}
}

@inproceedings{balin2019concrete,
  title={Concrete autoencoders: Differentiable feature selection and reconstruction},
  author={Bal{\i}n, Muhammed Fatih and Abid, Abubakar and Zou, James},
  booktitle={International conference on machine learning},
  pages={444--453},
  year={2019},
  organization={PMLR}
}

@article{mitchell2020global,
  title={Global ageing: successes, challenges and opportunities},
  author={Mitchell, Emma and Walker, Richard},
  journal={British journal of hospital medicine},
  volume={81},
  number={2},
  pages={1--9},
  year={2020},
  publisher={MA Healthcare London}
}

@article{almukadi2024deep,
  title={Deep feature fusion with computer vision driven fall detection approach for enhanced assisted living safety},
  author={Almukadi, Wafa Sulaiman and Alrowais, Fadwa and Saeed, Muhammad Kashif and Yahya, Abdulsamad Ebrahim and Mahmud, Ahmed and Marzouk, Radwa},
  journal={Scientific Reports},
  volume={14},
  number={1},
  pages={21537},
  year={2024},
  publisher={Nature Publishing Group UK London}
}

@inproceedings{li2009accurate,
  title={Accurate, Fast Fall Detection Using Gyroscopes and Accelerometer-Derived Posture Information},
  author={Li, Qiang and Stankovic, John A. and Hanson, Mark A. and Barth, Adam T. and Lach, John and Zhou, Gang},
  booktitle={Proceedings of the 2009 Sixth International Workshop on Wearable and Implantable Body Sensor Networks},
  year={2009},
  organization={IEEE},
  doi={10.1109/BSN.2009.46}
}

@article{mastorakis2014fall,
  title={Fall detection system using Kinect's infrared sensor},
  author={Mastorakis, Georgios and Makris, Dimitrios},
  journal={Journal of Real-Time Image Processing},
  volume={9},
  number={4},
  pages={635--646},
  year={2014},
  publisher={Springer},
  doi={10.1007/s11554-012-0246-9}
}

@article{li2023novel,
  title={Novel Fall Detection Algorithm based on Multi-Threshold Fall Model},
  author={Li, Hao and Ma, Jun and Ren, Xunhuan and Wang, Kaiyu},
  year={2023},
  publisher={elib.bsu.by}
}

@article{lim2022application,
  title={The application of artificial intelligence and custom algorithms with inertial wearable devices for gait analysis and detection of gait-altering pathologies in adults: A scoping review of literature},
  author={Lim, Ashley Cha Yin and Natarajan, Pragadesh and Fonseka, R Dineth and Maharaj, Monish and Mobbs, Ralph J},
  journal={Digital Health},
  volume={8},
  pages={20552076221074128},
  year={2022},
  publisher={SAGE Publications},
  doi={10.1177/20552076221074128}
}

@article{singh2020sensor,
  title={Sensor technologies for fall detection systems: A review},
  author={Singh, Anuradha and Rehman, Saeed Ur and Yongchareon, Sira and Chong, Peter Han Joo},
  journal={IEEE Sensors Journal},
  volume={20},
  number={13},
  pages={6889--6919},
  year={2020},
  publisher={IEEE},
  doi={10.1109/JSEN.2020.2975522}
}

@article{miranda2022survey,
  title={A survey on the use of machine learning methods in context-aware middlewares for human activity recognition},
  author={Miranda, Leandro and Viterbo, José and Bernardini, Flávia},
  journal={Artificial Intelligence Review},
  volume={55},
  number={4},
  pages={3369--3400},
  year={2022},
  publisher={Springer},
  doi={10.1007/s10462-021-10093-5}
}

@inproceedings{turetta2025lightweight,
  title={A Lightweight CNN for Real-Time Pre-Impact Fall Detection},
  author={Turetta, Cristian and Ali, Muhammad Toqeer and Demrozi, Florenc and Pravadelli, Graziano},
  booktitle={2025 Design, Automation \& Test in Europe Conference (DATE)},
  pages={1--7},
  year={2025},
  organization={IEEE}
}

@article{qu2024physics,
  title={Physics Sensor Based Deep Learning Fall Detection System},
  author={Qu, Zeyuan and Huang, Tiange and Ji, Yuxin and Li, Yongjun},
  journal={arXiv preprint arXiv:2403.06994},
  year={2024}
}

@article{mir2025machine,
  title={Machine Learning in Ambient Assisted Living for Enhanced Elderly Healthcare: A Systematic Literature Review},
  author={Mir, Aabid A and Khalid, Ahmad S and Musa, Shahrulniza and Fauzi, Mohammad Faizal Ahmad and Razak, Normy N and Tang, Tong Boon},
  journal={IEEE Access},
  year={2025},
  publisher={IEEE}
}

@article{li2025multidimensional,
  title={Multidimensional Time Series Segmentation of Human Activity Without Prior Knowledge},
  author={Li, Ping and Jiang, Ming and Lin, Huipin and Lv, Xudong and Huang, Jiye},
  journal={IEEE Internet of Things Journal},
  year={2025},
  publisher={IEEE}
}

@article{de2022fallCombinedDisplacement,
  title={Fall detection approach based on combined displacement of spatial features for intelligent indoor surveillance},
  author={De, Anurag and Saha, Ashim and Kumar, Praveen},
  journal={Multimedia Tools and Applications},
  volume={81},
  number={4},
  pages={5113--5136},
  year={2022},
  publisher={Springer},
  doi={10.1007/s11042-021-11646-w},
  url={https://doi.org/10.1007/s11042-021-11646-w}
}

@article{de2022fallSpatioTemporalFusion,
  title={Fall detection method based on spatio-temporal feature fusion using combined two-channel classification},
  author={De, Anurag and Saha, Ashim and Kumar, Praveen and Pal, Gautam},
  journal={Multimedia Tools and Applications},
  volume={81},
  number={18},
  pages={26081--26100},
  year={2022},
  publisher={Springer},
  doi={10.1007/s11042-022-11914-3},
  url={https://doi.org/10.1007/s11042-022-11914-3}
}
% \bibliography{sn-bibliography}% common bib file

\end{document}